\documentclass[final,3p,times,twocolumn]{elsarticle}

\usepackage{amssymb}
\usepackage{amsmath}
\usepackage{hyperref}
\usepackage{orcidlink} 
\usepackage{comment}
\usepackage[most]{tcolorbox}
\usepackage[utf8]{inputenc}
\usepackage{amsfonts}     
\usepackage{graphicx}     
\usepackage{booktabs}     

\usepackage{stfloats}
\usepackage{bm} 

\newcommand{\msd}[2]{\ensuremath{#1_{\small \pm #2}}}
\newcommand{\bmsd}[2]{\ensuremath{\bm{#1_{\small \pm #2}}}}

\usepackage{graphicx}
\usepackage{booktabs}
\usepackage{adjustbox}

\journal{Medical Image Analysis}

\begin{document}

\begin{frontmatter}

\title{Multi-FRuGaL: Multimodal Flexible Redundancy-aware Decomposed Gated Learning for Cancer Diagnosis and Prognosis}



\author[label1,label2]{Sanket Kachole%
  \href{https://orcid.org/0000-0002-1496-2070}{\orcidlink{0000-0002-1496-2070}}}

\author[label1,label2]{Siddhesh Thakur%
  \href{https://orcid.org/0000-0003-4807-2495}{\orcidlink{0000-0003-4807-2495}}}

\author[label1,label2]{Shubham Innani%
  \href{https://orcid.org/0000-0003-3616-0308}{\orcidlink{0000-0003-3616-0308}}}

\author[label1,label2]{Sanyukta Adap%
  \href{https://orcid.org/0009-0006-3639-0911}{\orcidlink{0009-0006-3639-0911}}}

\author[label1,label2]{Suhang You%
  \href{https://orcid.org/0000-0002-9327-9391}{\orcidlink{0000-0002-9327-9391}}}

\author[label1,label2]{Carla Pitarch-Abaigar%
  \href{https://orcid.org/0000-0002-6015-244X}{\orcidlink{0000-0002-6015-244X}}}

\author[label1,label2,label3,label4]{Spyridon Bakas\corref{cor}%
  \href{https://orcid.org/0000-0001-8734-6482}{\orcidlink{0000-0001-8734-6482}}}


\cortext[cor]{Corresponding author: spbakas@iu.edu}

\affiliation[label1]{
  organization={Division of Computational Pathology, Department of Pathology and Laboratory Medicine}, 
  addressline={Indiana University School of Medicine}, 
  city={Indianapolis}, 
  state={IN}, 
  country={USA}
}

\affiliation[label2]{
  organization={IU Melvin and Bren Simon Comprehensive Cancer Center}, 
  city={Indianapolis}, 
  state={IN}, 
  country={USA}
}

\affiliation[label3]{
  organization={Departments of Biostatistics and Health Data Science; Radiology and Imaging Sciences; Neurological Surgery; Indiana University School of Medicine}, 
  city={Indianapolis}, 
  state={IN}, 
  country={USA}
}
\affiliation[label4]{
  organization={Department of Computer Science, Luddy School of Informatics, Computing, and Engineering}, 
  city={Indianapolis}, 
  state={IN}, 
  country={USA}
}

\begin{abstract}
Modern medicine relies on heterogeneous data sources spanning radiology, pathology, text reports, and structured clinical information. However, real-world patient data are frequently incomplete, with missing or sparsely acquired modalities, limiting the effectiveness of standard multimodal fusion approaches. To this end, we propose the \textbf{M}ultimodal \textbf{F}lexible \textbf{R}ed\textbf{u}ndancy-aware decomposed \textbf{GA}ted \textbf{L}earning (Multi-FRuGaL) framework, a decomposition-aware, adaptive gated intermediate-fusion framework that performs modality-level representation learning under missing data. Multi-FRuGaL integrates per-modality encoders with a signal decomposition layer, an input-conditioned gating network, and an information-aware fusion objective to separate redundant from modality-specific complementary signals, selectively upweighting informative modalities and suppressing redundant or noisy inputs, and remaining well-defined even when multiple modalities are absent. We evaluate Multi-FRuGaL on two multimodal head and neck cancer cohorts: the HANCOCK challenge dataset ($N=763$) comprising five modalities and two prognostic endpoints (5-year survival and 2-year recurrence), and the HECKTOR challenge dataset ($N=588$) comprising three modalities for human papillomavirus (HPV) status classification. Multi-FRuGaL consistently achieves higher mean performance than the evaluated baselines across multiple tasks, improving AUC from $0.601$ to $0.8496$ for survival, from $0.672$ to $0.8102$ for recurrence, and achieving $0.975$ AUC for HPV prediction on HECKTOR. For survival analysis, it further achieves a concordance index of $0.6814$ for overall survival, $0.7421$ for recurrence-free survival, and $0.7143$ for progression-free survival on HANCOCK, and $0.7203$ for recurrence-free survival on HECKTOR. Qualitative analyses further show that Multi-FRuGaL learns discriminative and robust multimodal representations, even under severe missing-modality conditions. Source code implementation is available in \href{https://github.com/IUCompPath/Multi_Frugal/}{GitHub}.
\end{abstract}

\begin{keyword}
    Multimodal, Missing modalities, Pathology, HNSCC, Reports, TMA, Tabular
\end{keyword}

\end{frontmatter}



\section{Introduction}
\label{sec:intro}
    Modern medicine is undergoing a rapid digital transformation, with data-centric methodologies increasingly shaping clinical decision-making and patient care \cite{shilo2020axes}. Emerging technologies such as digital pathology \cite{niazi2019digital, srinidhi2021deep, chandrasekaran2026pgcgan}, biosensors \cite{sempionatto2022biosensors}, advanced imaging \cite{castiglioni2021artificial, kachole2024asynchronous}, and next-generation sequencing \cite{steyaert2023integrative, rehman2026lifetime} are generating unprecedented volumes of heterogeneous healthcare data across modalities, including radiology, pathology, -omics, and clinical categorical data, each providing complementary insights into patient health \cite{wang2024deep}. Traditionally, clinical decision-making has been based on multidisciplinary boards, where specialists interpret their respective modalities independently and collectively determine treatment strategies \cite{rosenblum2022multidisciplinary}. However, such human-driven processes are inherently limited in scalability, consistency, and specialized expertise \cite{rajpurkar2022ai}. Recent advances in artificial intelligence (AI) have begun to address these challenges through multimodal learning frameworks that can jointly analyze diverse data types and reason over their complementary information \cite{li2024multimodal, kachole2024bimodal}. These multimodal AI methods are typically trained to analyze specific heterogeneous sources of data in tandem, leveraging cross-modal correlations to form a holistic understanding of the patient’s condition. Despite this progress, the field remains constrained by challenges related to the vastly heterogeneous modality characteristics, inconsistent data availability, and the lack of robust fusion mechanisms that can distinguish common from modality-specific signals under incomplete-modality conditions \cite{schouten2025scoping}.

    While multimodal integration \cite{lipkova2022artificial} holds clear promise for precision oncology, existing fusion strategies remain limited in robustness and depth \cite{huang2020fusion, castiglioni2021artificial, rajpurkar2022ai}. Among early, intermediate, and late multimodal fusion strategies, the latter remains the most common approach. However, late fusion relies on independent, unimodal predictions, thereby failing to capture the complex, non-linear interactions across different modalities. In contrast, intermediate fusion enables joint representation learning within a shared latent space, allowing the model to synthesize complementary features from each modality before the final decision stage \cite{baltruvsaitis2019multimodal, li2023adaptivefusion}. Nevertheless, most previous frameworks employ simple concatenation or attention-based fusion across a small number of modality-derived features, typically combining clinical, radiomic, and histopathologic features, yet they assume that all modalities are simultaneously available at inference \cite{chen2020pathomic, soenksen2022integrated}. In practice, real-world clinical datasets are characterized by asymmetric modality availability, with missing or incomplete data arising from non-standardized imaging protocols, variable tissue sampling, and heterogeneous documentation quality \cite{suter2023missing, zhang2021missing}. Conventional imputation-based approaches \cite{chen2025alignment, boyko2025imputmae} often propagate noise or bias when reconstructing these missing modalities, leading to unstable or uninterpretable methods \cite{perez2023handling, zhao2023moddrop}. Moreover, common fusion schemes rarely account for information redundancy, allowing overlapping cross-modal signals to be repeatedly encoded and potentially dominate the fused representation \cite{baltruvsaitis2019multimodal, tsai2019multimodal}. In addition, existing fusion methods typically do not explicitly distinguish shared/redundant disease-relevant information across modalities from modality-specific signals, making it difficult to determine whether performance gains arise from complementary evidence or from duplicated cross-modal information. Addressing these challenges requires a principled framework that can dynamically weight, gate, or suppress modality contributions, as well as decompose common and modality-specific signals based on their informativeness, enabling adaptive, interpretable, and redundancy-aware fusion under high missingness \cite{li2023adaptivefusion, wang2024selfsupervised}.

    To address these methodological challenges, we propose the \emph{\textbf{Multi}modal \textbf{F}lexible \textbf{R}ed\textbf{u}ndancy-aware decomposed \textbf{Ga}ted Learning} (Multi-FRuGaL), an intermediate-fusion framework designed to perform robust multimodal fusion learning, under high and irregular missingness. Multi-FRuGaL introduces a signal decomposition (SD) layer that separates each modality's representations into common and modality-specific components. It then applies an input-conditioned gating mechanism (GateNet) that adaptively allocates representational weight across modalities, which allows the framework to either amplify informative modalities or suppress noisy, redundant, or even entirely missing sources. The architecture is trained using (i) a task loss ($L_{\text{task}}$) based on binary cross-entropy for outcome prediction, and (ii) a decomposition loss ($L_{\text{dec}}$) that separates modality-shared and modality-specific signals using orthogonality constraint, and (iii) a Information-Budget Regularizer ($L_{\text{IBR}}$), an information-aware fusion loss that regularizes how modalities contribute during fusion. Specifically,  $L_{\text{IBR}}$ comprises (a) a redundancy penalty ($L_{\text{red}}$), which minimizes overlapping information to prevent collinear signals from dominating the fusion space, and (b) a sparsity-inducing budget term ($L_{\text{budget}}$), which applies an $\ell_{1}$-style penalty on gate activations to promote efficient and selective modality usage. Unlike conventional fusion approaches that implicitly treat all modalities as equally valuable \cite{baltruvsaitis2019multimodal,tsai2019multimodal}, Multi-FRuGaL provides an explicit mechanism for information allocation across heterogeneous and incomplete data sources. We independently evaluate the proposed framework on survival, recurrence, and human papillomavirus (HPV)-status prediction tasks from varying head and neck cancer multimodal datasets, spanning across the MICCAI 2025 HANCOCK \cite{dorrich2025multimodal} and the MICCAI 2025 HECKTOR \cite{andrearczyk2023overview} benchmarking environments. Importantly, Multi-FRuGaL is evaluated across different imaging data streams—with HECKTOR providing neuroimaging and clinical tabular data, whereas HANCOTHON provides clinical tabular data, clinical reports, and pathology whole-slide images (WSI) coupled with tissue microarrays (TMA).

    The primary contributions of this study are 4-fold:
    \begin{enumerate}
        \item We introduce \textit{Multi-FRuGaL}, a novel intermediate-fusion framework that combines a signal decomposition layer and input-conditioned stochastic gating with a predefined information budget to enable adaptive modality routing and robust modality fusion at high rates of missing data, without requiring imputation.
        \item We introduce a mask-aware pooling mechanism that defines a well-posed aggregation operator under partial modality availability, ensuring missing modalities do not contribute to the fused patient representation.
        \item We introduce a decomposition-aware training objective that combines a decomposition loss ($L_{\text{dec}}$) for separating modality-shared and modality-specific signals with the Information-Budget Regularizer ($L_{\text{IBR}}$), that couples a sparsity-inducing budget penalty ($L_{\text{budget}}$) and redundancy penalty ($L_{\text{red}}$), to guide the underlying fusion framework to select the most complementary and non-redundant information.
        \item We offer real-world data validation on two independent clinical tasks across distinct benchmark environments (HANCOTHON \& HECKTOR), demonstrating that Multi-FRuGaL outperforms standard fusion baselines in high missingness settings. We also show that the learned gates provide per-patient ``information allocation profiles''.  
    \end{enumerate} 


\section{Related Work}
\label{section:Related Work}
    \subsection{Multimodal Fusion Architectures in Oncology}
        In multimodal learning, the integration of heterogeneous data sources can occur at different stages of the method pipeline, commonly referred to as early, intermediate, and late fusion \cite{lipkova2022artificial}. In early fusion, modalities are combined prior to feature encoding, enabling joint optimization of feature representations but requiring the inputs to reside in the same representational space \cite{barnum2020benefits}. Late fusion, by contrast, combines the outputs of separate unimodal methods at the decision level, which facilitates flexibility in handling unpaired data but precludes interaction between modalities during training \cite{nikolaou2025machine}. Intermediate fusion, where modalities are fused after feature encoding but before the final task-specific layers, shows the most promise as an effective strategy. This approach enables the method to capture complementary information across modalities while preserving modality-specific representations, resulting in improved performance compared to early and late fusion \cite{guarrasi2025systematic, kachole2024bimodal, huang2024neuromorphic}.

        Recent research trends showcase a rise in intermediate fusion architectures and training methods that emphasize fine-grained cross-modal alignment, moving beyond classical contrastive objectives \cite{ramanathan2025modaltune, ramanathan2024ensemble, chen2022head, naeini2022event, kachole2023asynchronous, kachole2020computer}. Methods such as COSMOS \cite{kim2024cosmos} leverage cross-attention-based cross-modal self-distillation to align local and global augmentations, while medical frameworks (e.g., MIMO \cite{chen2025mimo}) combine textual and visual features within a shared encoder–decoder pipeline to produce pixel-grounded text outputs. In parallel, another line of work addresses the unique challenges of vision–tabular reasoning: SynTab-LLaVA \cite{zhou2025syntab} employs decoupled data synthesis to enable relational understanding between image and structured data, whereas STiL \cite{du2025stil} learns disentangled modality-shared and modality-specific representations, representing a canonical intermediate fusion framework for semi-supervised settings. As multimodal methods scale, a third trend tackles hallucination suppression through fusion regularization. ClearSight \cite{yin2025clearsight} strengthens visual grounding via an additive amplification module within vision–language transformers; while not a distinct fusion type, it enhances intermediate fusion by reweighting attention. Meanwhile, Nullu \cite{yang2025nullu} performs post-hoc null-space projection, editing method weights after training to remove latent directions associated with hallucinated outputs. Finally, VLU-Net \cite{zeng2025vlu} explores dynamic intermediate fusion, integrating vision and language through iterative conditioning within a deep unfolding architecture, thereby achieving cross-modal interactions in the latent space. Collectively, these advances demonstrate a shift from static fusion toward adaptive, semantically grounded, and self-regularizing paradigms \cite{kachole2025posture, kachole2026object}. However, most still presuppose the availability of all modalities, leaving the problem of robust imputation and reweighting under real-world missing-modality conditions largely unresolved.

    \subsection{Techniques Handling Missing Modalities}
        Before the emergence of deep generative approaches, conventional imputation strategies such as mean or zero-filling, $k$-nearest neighbors (KNN) imputation \cite{troyanskaya2001missing}, and matrix factorization–based techniques \cite{stekhoven2012missforest,josse2016missmda} were widely employed to address missing data in radiomics and -omics pipelines. Although computationally efficient, these methods rely on simplistic distributional assumptions and fail to capture nonlinear inter-modal dependencies, often introducing bias when modality-level information is absent.

        To overcome these limitations, recent multimodal oncology research has converged on three main strategies. The first, generative imputation, reconstructs missing modalities by learning inter-modal dependencies within a shared latent space. Methods such as impuTMAE \cite{boyko2025imputmae} employ masked transformer pretraining to predict missing channels from observed ones, while variational frameworks like Multimodal CustOmics \cite{benkirane2023customics, you2025profuseme} and related VAE-based designs synthesize incomplete omics data through latent-space sampling. Similarly, methods such as Pathomic Fusion \cite{chen2020pathomic} and Wang et al. \cite{wang2025missing} extend this principle by jointly encoding available modalities and inferring proxy embeddings for absent streams, thereby preserving downstream predictive capacity. 

        A second, complementary direction emphasizes architectural robustness to missing inputs. Rather than reconstructing absent data, these frameworks ensure that fusion remains stable even when one or more modalities are unavailable. This robustness is often achieved through *modality dropout*—randomly masking input streams during training to prevent over-reliance on any single source—or through adaptive gating and cross-attention mechanisms, as demonstrated by Cui et al. \cite{cui2022multi,cui2023deep}, which dynamically reweights available features. Temporal and longitudinal architectures, such as Yeghaian et al. \cite{yeghaian2025multimodal}, extend this paradigm to partially observed time-series data, enabling consistent predictions under irregular sampling. 

        Finally, a third trend applies knowledge distillation and representation transfer to bridge complete and incomplete modalities. In this setup, a student model trained on partial inputs learns to emulate a teacher network exposed to the full modality set. The Distilled Prompt Learning framework \cite{xu2025distilled} exemplifies this approach, leveraging prompt-based conditioning and cross-modal supervision to achieve robust survival prediction even when modalities are missing. Collectively, these developments represent a shift from rigid multimodal dependence toward flexible, self-correcting fusion systems capable of maintaining performance under real-world data missingness.

        While the strategies of generative imputation, architectural robustness, and knowledge distillation have each advanced multimodal learning under data missingness, they remain fundamentally limited. Generative methods risk error propagation when imputations deviate from biological reality; robust architectures, by design, cannot recover entirely absent information; and distillation-based methods inherit the biases and constraints of their teacher networks, requiring complete data during training. Moreover, existing frameworks typically assume that all modalities are equally informative and available, lacking explicit mechanisms to balance redundancy, information efficiency, or reliability under partial observability~\cite{pooja2022redundancy}. Although recent advances such as modality dropout and self-supervised alignment have improved empirical robustness~\cite{suter2023missing,wang2024selfsupervised}, these techniques remain heuristic and do not provide any systematic way to determine how much, or even when, each modality should contribute. To address these limitations, we introduce the Multi-FRuGaL framework that unifies adaptive gating with an information-budget regularizer, enabling redundancy-aware and dynamically weighted multimodal fusion under arbitrary missingness.


\section{Methodology}
\label{section:Methodology}
\begin{figure*}[t]
            \centering
            \setlength{\tabcolsep}{0pt}
            \setlength{\arrayrulewidth}{0.5pt}
            \begin{tabular}{|c|}
                \hline
                \hspace*{-0.10\linewidth}%
                \includegraphics[
                    width=0.7\linewidth,
                    height=0.28\textheight,
                    keepaspectratio
                ]{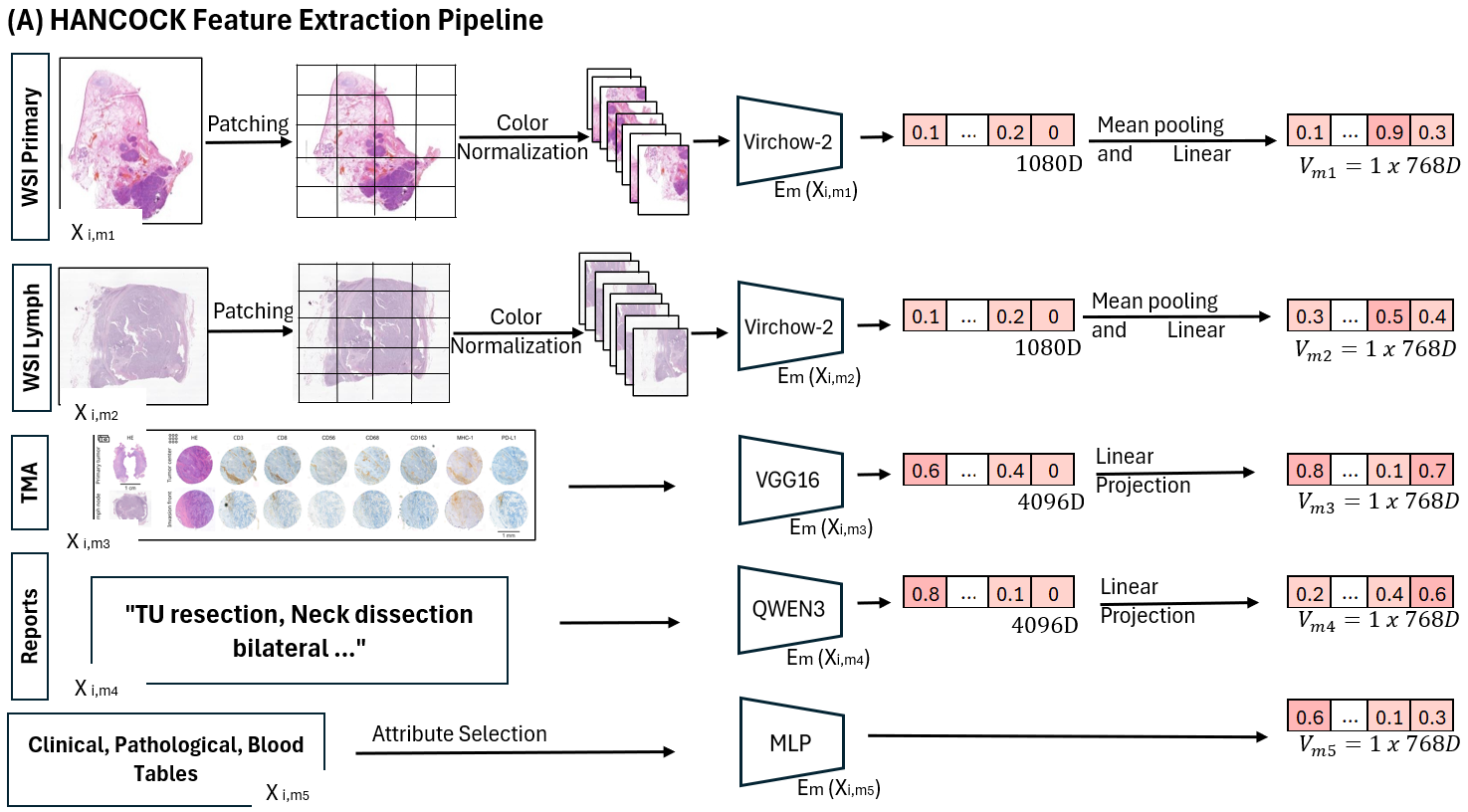} \\[0.3em]
                \hline
        
                \hspace*{-0.09\linewidth}%
                \includegraphics[
                    width=0.7\linewidth,
                    height=0.23\textheight,
                    keepaspectratio
                ]{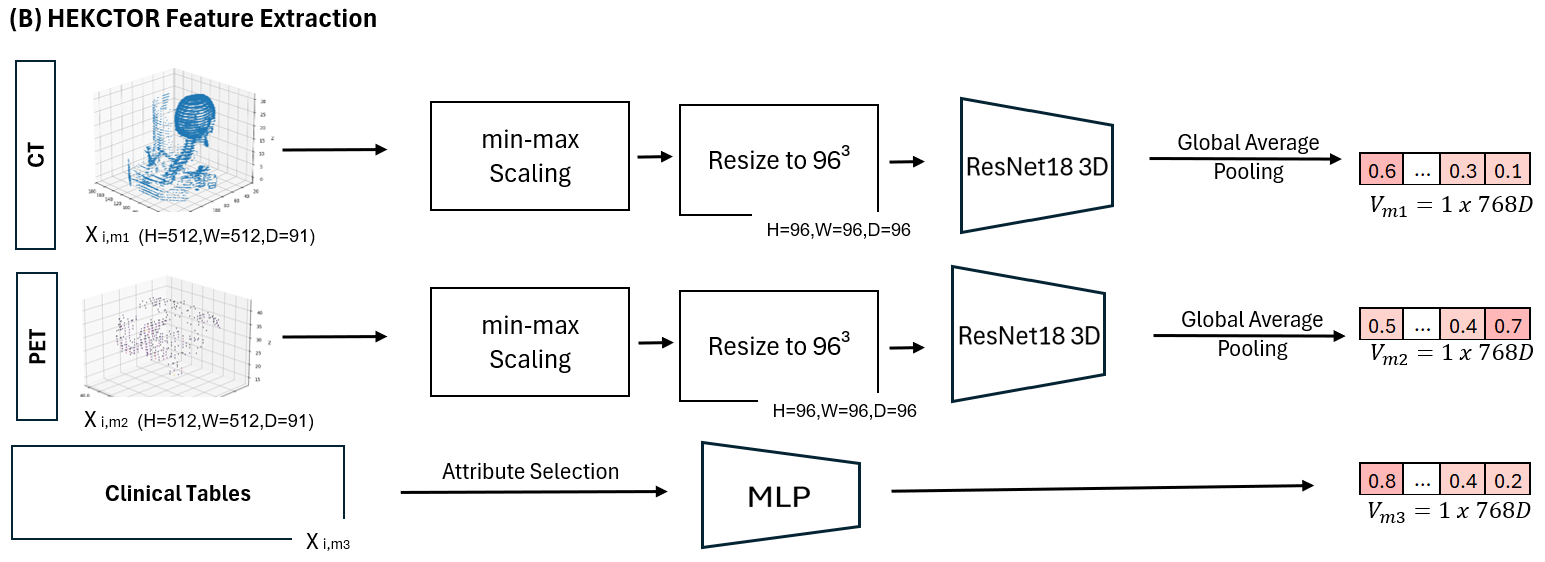} \\[0.3em]
                \hline
        
                \includegraphics[
                    width=0.8\linewidth,
                    height=0.49\textheight,
                    keepaspectratio
                ]{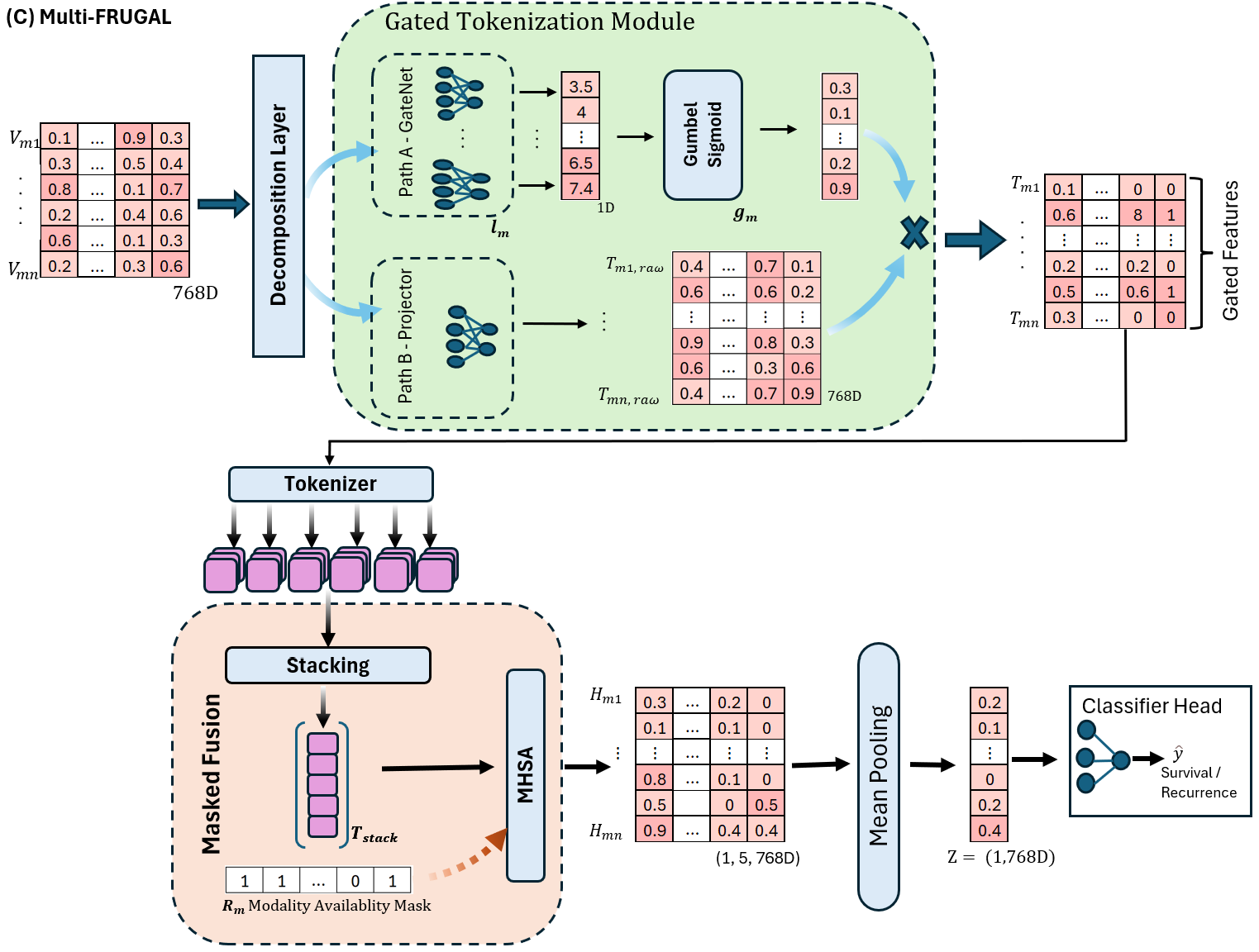} \\[0.3em]
                \hline
            \end{tabular}
            \caption{
                Overview of the proposed Multimodal Flexible Redundancy-aware decomposed Gated Learning (Multi-FRuGaL) framework.
                \textbf{(A)} Data processing and feature extraction pipeline for the HANCOCK dataset.
                \textbf{(B)} Data processing and feature extraction pipeline for the HECKTOR dataset.
                \textbf{(C)} Multi-FRuGaL framework illustrating gated fusion, tokenization, and transformer-based multimodal aggregation for downstream classification.
            }
            \label{fig:Multi-FRUGAL_overview}
        \end{figure*}

    This section details our Multi-FRuGaL framework (Fig.~\ref{fig:Multi-FRUGAL_overview}). We first describe the unimodal feature extraction process (\ref{subsection:Unimodal Feature Extraction}). We then introduce the core fusion module (\ref{section:information_budgeted_fusion}), which integrates these features using input-conditioned gating and a masked Transformer. Finally, we define the joint objective function (\ref{subsection:Objective_Function}) that guides the framework, combining a supervised task loss ($L_{\text{task}}$), a decomposition loss ($L_{\text{dec}}$), and an information-budget regularizer ($L_{\text{IBR}}$).

    \subsection{Problem Formulation}
    \label{subsection:Problem Formulation}
        Let a patient cohort be represented by a dataset $\mathcal{D} = \{(\mathcal{X}_i, \mathbf{R}_i, y_i)\}_{i=1}^N$, where $N$ is the total number of patients. Each patient $i$ is described by a set of $M$ heterogeneous modalities, $\mathcal{X}_i = \{X_{i,1}, \ldots, X_{i,M}\}$. The number and type of modalities ($M$) may vary across datasets but include histopathology (e.g., WSI, TMA), radiology, genomics, free-text reports, and structured tabular data. A key challenge in real-world clinical settings is data missingness. We define a binary modality-availability mask $\mathbf{R}_i \in \{0, 1\}^M$, where $R_{i,m} = 0$ if modality $X_{i,m}$ is absent for patient $i$, and $R_{i,m} = 1$ otherwise. This formulation explicitly models the arbitrary missingness patterns characteristic of clinical cohorts. The target variable $y_i \in \mathcal{Y}$ represents the task-specific binary clinical endpoint, with $\mathcal{Y}=\{0,1\}$. Dataset-specific definitions of these endpoints are provided in the experimental evaluation section \ref{subsection:Experiments Evaluation}.

        The objective is to learn a single and generalizable predictive function $f_\theta$ that maps an incomplete multimodal set to a prediction $\hat{y}_i$:

        \begin{equation}
            \hat{y}_i = f_\theta(\mathcal{X}_i, \mathbf{R}_i),
        \end{equation}

        \noindent where $f_\theta$ is trained to optimally exploit the available information in $\mathcal{X}_i$ while remaining inherently resilient to diverse missingness patterns specified by $\mathbf{R}_i$.

    \subsection{Feature Extraction}
    \label{subsection:Unimodal Feature Extraction}
        Each raw modality $X_m$ undergoes modality-specific pre-processing and feature extraction prior to fusion, as illustrated in Fig.~\ref{fig:Multi-FRUGAL_overview}(A-B). The extraction process of individual modalities, including pathology, radiology, text reports, and tabular data, is described in subsequent subsections. 

        \subsubsection{Pathology Feature Extraction}
        \label{subsubsection:Pathology Feature Extraction}
            The extraction of features from image-based modalities is a multi-step process. For the gigapixel primary-tumor WSI and lymph-node WSIs, we first employ a robust tissue segmentation pipeline based on CLAM \cite{lu2021data} to distinguish foreground tissue from the white background. These tissue regions are then partitioned into a sequence of $20\times$ magnification (0.5mpp) patches. Self-supervised representation learning has shown strong potential for improving digital histopathology feature extraction from whole-slide image patches~\cite{ciga2022self}. Patch-level embeddings are then extracted using the pre-trained foundation pathology encoder, Virchow2 \cite{zimmermann2024virchow2}, which was selected due to its large-scale pretraining on diverse histopathology data and prior literature \cite{neidlinger2025benchmarking} showing strong performance relative to earlier encoders. Virchow2 produced a patch-embedding matrix $F_{m} \in \mathbb{R}^{P_m \times d_m}$ for each slide, where $P_m$ is the number of patches and $d_m$ is the feature dimension. A mean-pooling operation converts these patch embeddings into compact patient-level vectors $V_m \in \mathbb{R}^{1 \times 1280}$. For the tissue microarray (TMA) modality, we directly process the provided TMA core images (available as pre-extracted PNG files), rather than gigapixel whole-slide images. Each core is encoded using a ImageNet-pretrained VGG16 as a classical baseline convolutional backbone (VGG16) \cite{simonyan2014very} to capture immunohistochemical texture and staining intensity, yielding a single 4096-dimensional feature vector per patient.
            
        \subsubsection{Radiology Feature Extraction}
        \label{subsubsection:Radiology Feature Extraction}
            Volumetric radiological features are extracted independently from the contrast-enhanced CT and the FDG-PET scans using parallel 3D ResNet-18 encoders~\cite{hara2018can}. For each patient, the original CT volumes ($512 \times 512 \times 91$) and FDG-PET volumes ($128 \times 128 \times 91$) are first converted to a channel-first representation and then intensity-normalized using per-volume min–max scaling. These normalized volumes are resized to a fixed spatial resolution of $96 \times 96 \times 96$, to enable batch-wise processing and architectural consistency across patients and modalities. Each modality is processed by a dedicated 3D ResNet-18 backbone with a one-input channel, since CT and PET are represented as single-channel volumetric intensity images. The final classification layer of each network is removed, and global average pooling is applied over the spatial dimensions of the final 3D feature map to obtain one 512-dimensional vector per modality. This results in separate 512-dimensional feature vectors for CT and FDG-PET for each patient. The extracted radiological embeddings are stored for downstream multimodal fusion and predictive modeling.

        \subsubsection{Text Reports Feature Encodings}
        \label{subsubsection:Text Reports Feature Encodings}
            The free-text modality is encoded into a patient-level feature vector $V_m \in \mathbb{R}^{1 \times d_m}$ using the comprehensive \texttt{TextData} repository within the HANCOCK dataset \cite{doerrich2024multimodal}. This repository contains multiple narrative sources, including \texttt{histories}, \texttt{reports}, and \texttt{surgery\_descriptions}, each provided in both German and English. For framework consistency and to leverage publicly available encoders, we utilized only the English-translated versions. For each patient, the texts from \texttt{histories\_english}, \texttt{reports\_english}, and \texttt{surgery\_descriptions\_english} were concatenated into a single, comprehensive document. The semi-structured \texttt{icd\_codes} and \texttt{ops\_codes} were omitted from this modality as their content is better captured by the structured tabular data explained in the next section \ref{subsubsection:Tabular Feature Encodings}. This consolidated document is processed using the Qwen3-8B large language model \cite{bai2023qwen}. We extract the pooled output embedding from the final layer to produce a high-dimensional, contextualized embedding of dimension $d_m = 4,096$, which serves as the unimodal representation $V_m$. 

        \subsubsection{Tabular Feature Encodings}
        \label{subsubsection:Tabular Feature Encodings}
            The structured tabular data source is a composite of three distinct, patient-aligned clinical datasets, including demographics and age related information, pathological data, including tumor primary site, staging, margins information, and blood data, including lab values. After merging these sources, we performed attribute selection based on the provided data dictionaries and clinical relevance. Preprocessing was data-type-specific. Categorical attributes were one-hot encoded. Numerical attributes were Z-score normalized. Crucially, we performed clinical feature engineering on the \texttt{blood\_data}. Using the provided \texttt{blood\_data\_reference\_ranges}, we created new binary features to flag lab values as ``Normal'' or ``Abnormal''. We also computed clinically-relevant derived features, such as the neutrophil-to-lymphocyte ratio (NLR). Finally, all processed numerical, categorical, and derived binary features were concatenated to form a single, patient-wise tabular feature vector $V_m \in \mathbb{R}^{1 \times d_m}$.

            The resulting patient-level representations from all modalities are denoted as

            \begin{equation}
                \{V_m\}_{m=1}^{M}, \quad V_m \in \mathbb{R}^{1 \times d_m},
            \end{equation}
            
            \noindent where each index $m$ corresponds to a distinct modality present in the dataset.

            To harmonize dimensionality across modalities, each embedding is passed through a  linear projector
            \[
            \tilde V_m = P_m V_m, \quad P_m \in \mathbb{R}^{d \times d_m},
            \]
            followed by LayerNorm, yielding tokens in a shared latent space $\tilde V_m \in \mathbb{R}^{1 \times 768}$. 
            These projected tokens serve as the input to the Multi-FRuGaL Fusion Module for joint representation learning.


    \subsection{Multimodal Flexible Redundancy-aware decomposed Gated Learning}
    \label{section:information_budgeted_fusion}
        The Multi-FRuGaL framework adaptively fuses a collection of $M$ modality-specific feature vectors $\{\tilde V_m\}_{m=1}^{M}$, where each $\tilde V_m \in \mathbb{R}^{1 \times d}$ . As illustrated in Fig.~\ref{fig:Multi-FRUGAL_overview}(B), the module consists of four stages: (i) a signal decomposition layer (ii) a gated tokenization module, (iii) a masked fusion encoder, and (iv) a prediction head.
 
        \subsubsection{Signal Decomposition Layer}
        \label{subsubsection:Signal_Decomposition_Layer}
            Each modality-specific embedding is first decomposed into modality-shared and modality-specific components in the common latent space. Let $\tilde V_m \in \mathbb{R}^{d}$ denote the projected representation of modality $m$. Rather than relying on a single latent representation per modality, the proposed decomposition layer factorizes $\tilde V_m$ into modality-shared and modality-specific components using two modality-specific projection heads:
            \begin{equation}
            s_m = f_m^{(s)}(\tilde V_m), \qquad p_m = f_m^{(p)}(\tilde V_m),
            \end{equation}
            where $s_m \in \mathbb{R}^{d}$ represents the modality-shared component and $p_m \in \mathbb{R}^{d}$ represents the modality-specific component. The modality-shared branch is intended to capture disease-relevant information that is consistent across modalities, whereas the modality-specific branch preserves information unique to each modality. The separation between these two components is enforced through an orthogonality-based decomposition, implemented with a squared cosine penalty, which encourages them to encode distinct, non-overlapping information prior to fusion. The decomposed components are then concatenated to preserve their distinct information content:
            \begin{equation}
            u_m = [\, s_m \,\|\, p_m \,] \in \mathbb{R}^{2d}.
            \end{equation}

        \subsubsection{Gated Tokenization Module}
        \label{subsubsection:Gated_Tokenization_Module}
        
            This module transforms each unimodal feature vector $\tilde V_m$ into a gated token $T_m \in \mathbb{R}^{768}$. The computation proceeds through two parallel paths, \textit{Path A} and \textit{Path B}, followed by a modulation step that applies the learned gate to the token. Let $R_m \in \{0,1\}$ denote the availability indicator for modality $m$, where $R_m=1$ and $R_m=0$ indicates that the modality is observed or unavailable, respectively. All modalities are kept in fixed slots; unavailable modalities are zeroed during gated modulation.

            \paragraph{\textit{Path A - Router}: GateNet}
                For each modality $m$, the gating network $\text{GateNet}_m$ receives the unimodal feature vector $\tilde V_m \in \mathbb{R}^{768}$ and outputs a scalar gating logit:
                \begin{equation}
                \ell_m = \mathrm{MLP}_m(\tilde V_m),
                \end{equation}
                where $\mathrm{MLP}_m: \mathbb{R}^{768} \rightarrow \mathbb{R}^{1}$.
                
                The logit is transformed into a continuous gate value using a Gumbel--Sigmoid relaxation \cite{maddison2016concrete}:
                \begin{equation}
                g_m = \mathrm{GumbelSigmoid}(\ell_m), \qquad g_m \in (0,1),
                \end{equation}
                which serves as a soft importance weight for modality $m$.  
                This gate is used to scale the corresponding modality representation during gated modulation, enabling the model to attenuate or emphasize modalities in a differentiable manner.

            \paragraph{\textit{Path B}: Projector}
                In parallel, the concatenated representation $u_m \in \mathbb{R}^{2d}$ is passed through a modality-specific linear projection layer to produce the raw modality token
                    \begin{equation}
                    T_{m,\text{raw}} = \mathrm{Proj}_m(u_m) \in \mathbb{R}^{768},
                    \end{equation}
                    which serves as the modality representation prior to gated modulation.

            \paragraph{Gated Modulation}
                The outputs of the gating and projection paths are combined through element-wise multiplication, where the \emph{soft} gate $g_m \in (0,1)$ is broadcast over the token dimension:
                
                \begin{equation}
                T_m =  R_m \, g_m \cdot T_{m,\text{raw}}.
                \label{eq:token}
                \end{equation}
                Here, $T_{m,\text{raw}} \in \mathbb{R}^{768}$ denotes the projected modality token, and $g_m$ softly modulates its contribution, producing the gated token $T_m \in \mathbb{R}^{768}$.  
                For available modalities ($R_m=1$), this operation  adaptively emphasizes or attenuates the modality token, while unavailable modalities ($R_m=0$) are zeroed out before fusion.

        \subsubsection{Masked Fusion Encoder}
        \label{subsubsection:Masked_Fusion_Encoder}
            After gated modulation, we obtain a set of $M$ modality tokens $\{T_m\}_{m=1}^{M}$, where each token $T_m \in \mathbb{R}^{768}$. The masked fusion encoder integrates these tokens into context-aware representations through stacking and masked self-attention.
            
            First, the tokens are stacked along the sequence dimension to form the input tensor:
            \[
            T_{\text{stack}} = \mathrm{stack}(T_1,\ldots,T_M) \in \mathbb{R}^{M \times 768}.
            \]
            
            The stacked tokens are processed by a multi-layer Transformer encoder~\cite{NIPS2017_3f5ee243}, which applies multi-head self-attention (MHSA) across the modality dimension. To prevent missing modalities from participating in the attention computation, we construct a binary attention mask derived solely from the modality availability vector
            \[
            \mathbf{R} = (R_1, \ldots, R_M), \qquad R_m \in \{0,1\}.
            \]
            
            For each query–key pair, the attention operation is defined as
            \begin{equation}
            \mathrm{Attn}(Q,K,V) = \mathrm{softmax}\!\left(\frac{QK^{T}}{\sqrt{d_k}} + \text{Mask}(\mathbf{R})\right)V,
            \end{equation}
             where $\text{Mask}(\mathbf{R})$ is implemented as a key-padding mask that prevents available modality tokens from attending to unavailable modalities. Since unavailable modality tokens are already zeroed during gated modulation and again excluded during mask-aware pooling, their outputs do not contribute to the final fused representation.
            
            The Transformer outputs a sequence of context-aware modality representations
            \[
            H = \{h_1, \ldots, h_M\}, \qquad H \in \mathbb{R}^{M \times 768}.
            \]
            
        \subsubsection{Prediction Head}
        \label{subsubsection:Prediction_Head}
            The prediction head converts the sequence of context-aware modality tokens into a final scalar prediction. As shown in Fig.~\ref{fig:Multi-FRUGAL_overview}, the prediction head consists of two components: a mask-aware pooling step and a linear classification layer.
            
            The fusion encoder outputs a set of modality tokens $H = \{h_1, \ldots, h_M\}$, where each $h_m \in \mathbb{R}^{768}$. To obtain a fixed-length patient representation, a mask-aware mean pooling operation aggregates only the tokens corresponding to available modalities. This ensures that suppressed or missing modalities do not influence the pooled representation. The pooled vector is computed as
            
            \begin{equation}
                Z =
                \frac{\sum_{m=1}^{M} R_m\, h_m}{\max\left(\sum_{m=1}^{M} R_m, 1\right)},
                \quad \text{with } \sum_{m=1}^{M} R_m \geq 1.
            \label{eq:mask_aware_pooling}
            \end{equation}

            resulting in the fused patient embedding $Z \in \mathbb{R}^{768}$. Samples are required to have at least one modality observed.

            The vector $Z$ is then processed by a linear classification head consisting of a fully connected layer with weight matrix $W_c \in \mathbb{R}^{768 \times 1}$ and bias $b_c \in \mathbb{R}$. The purpose of this layer is to learn a weighted combination of the 768 fused features, producing a scalar logit that reflects their contribution to the prognostic task. Applying the sigmoid function yields the final predicted probability:
            \begin{equation}
                \hat{y} = \sigma(W_c Z + b_c),
            \end{equation}
            where $\hat{y} \in [0,1]$ represents the predicted survival or recurrence probability and is used to compute the task-specific loss $L_{\text{task}}$ during training.

    \subsection{Objective Function}
    \label{subsection:Objective_Function}
        The entire framework, from the unimodal encoders $E_m$ through the gating networks to the fusion module, is trained end-to-end by minimizing a composite loss. This objective consists of a task-specific loss that drives predictive performance, a decomposition loss that disentangles modality-shared and modality-specific representations, and an information-budget regularizer that constrains how modalities contribute to the fused representation.
        \begin{equation}
            L_{\text{total}} = L_{\text{task}} +  L_{\text{dec}} + L_{\text{IBR}}
        \end{equation}
        
        The term $L_{\text{task}}$ optimizes the framework for the downstream prediction objective, such as survival or recurrence prediction. The decomposition loss $L_{\text{dec}}$ regularizes the signal decomposition layer by encouraging the modality-shared and modality-specific components within each modality to encode distinct, non-overlapping information. The information-budget regularizer $L_{\text{IBR}}$ governs the behavior of the gating mechanism by encouraging efficient modality usage while discouraging redundant or unnecessary modality contributions. Together, these complementary losses ensure that the network achieves high predictive accuracy while learning disentangled, frugal, and modality-aware multimodal representations.

        \subsubsection{Task-Specific Loss ($L_{\text{task}}$)}
        \label{subsubsection:Task_Loss}
            The task-specific loss drives the network to produce accurate prognostic classifications. For the binary tasks, including 5-year survival and 2-year recurrence prediction, we use the standard binary cross-entropy (BCE) loss, which quantifies the discrepancy between the predicted probability $\hat{y}$ and the ground-truth binary label $y \in \{0,1\}$:
            \begin{equation}
                L_{\text{task}} = -\left[y \log (\hat{y}) + (1-y)\log(1-\hat{y})\right],
                \label{eq:bce_loss}
            \end{equation}
            This loss provides gradients to all upstream components through the fused representation $Z$, including the modality tokens $T_m$, and the gating policy. 

        \subsubsection{Decomposition Loss ($L_{\text{dec}}$)}
        \label{subsubsection:Decomposition_Loss }
            To encourage the signal decomposition layer to disentangle modality-shared and modality-specific information within each modality, we impose an orthogonality constraint between the corresponding latent components. Let $s_m \in \mathbb{R}^{d}$ and $p_m \in \mathbb{R}^{d}$ denote the modality-shared and modality-specific embeddings for modality $m$, respectively, and let $R_m \in \{0,1\}$ indicate modality availability. The decomposition loss is defined as:
            \begin{equation}
                L_{\text{dec}}
                =
                \lambda_{\text{dec}} \frac{1}{\sum_{m=1}^{M} R_m}
                \sum_{m=1}^{M} R_m \, \cos^2(s_m, p_m).
            \end{equation}
            
            The squared cosine term penalizes both positive and negative alignment between $s_m$ and $p_m$, with its minimum achieved when $\cos(s_m,p_m)=0$. This encourages the modality-shared and modality-specific components to become orthogonal rather than merely anti-correlated, promoting distinct and non-overlapping representations prior to multimodal fusion.

    \subsubsection{Information-Budget Regularizer ($L_{\text{IBR}}$)}
    \label{subsubsection:Information_Budget_Regularizer}
        Our IBR enforces the principle that effective multimodal fusion should prioritize informative modalities while suppressing redundant, collinear, or unnecessary input information. Rather than providing direct supervision, IBR influences the learning dynamics by contributing gradients during end-to-end optimization, thereby shaping the gating behavior toward sparse and complementary modality utilization. It consists of two components:
        \begin{equation}
            L_{\text{IBR}} = L_{\text{budget}} + L_{\text{red}}.
        \end{equation}
        The budget term penalizes excessive modality usage, while the redundancy term discourages overlap in the information contributed by different modalities.

        \paragraph{Budget Penalty Regularizer ($L_{\text{budget}}$)}
            The budget penalty discourages unnecessary modality usage by applying a cost to activating modality gates. For each modality $m$, the gating network produces a \emph{soft} gate value $g_m \in (0,1)$ via a Gumbel--Sigmoid relaxation, which is used to scale the corresponding modality token during the forward pass. The budget penalty regularizes the expected magnitude of these soft gate activations for available modalities:
            \begin{equation}
                L_{\text{budget}} = \gamma \sum_{m=1}^{M} R_m \, \mathbb{E}[g_m],
                \label{eq:budget}
            \end{equation}
            where $R_m \in \{0,1\}$ indicates modality availability and $\gamma$ controls the strength of the penalty. In practice, the expectation is approximated by averaging the sampled soft gate values over the mini-batch. This term imposes a constant downward pressure on gate activations, such that a modality is assigned a higher gate value only when its contribution sufficiently reduces the task loss $L_{\text{task}}$ to offset the imposed budget cost.

        \paragraph{Redundancy Penalty Regularizer ($L_{\text{red}}$)}
            Let $p_m \in \mathbb{R}^{d}$ denote the modality-specific component for modality $m$, and let $R_m \in \{0,1\}$ indicate modality availability. We define $N_{\mathrm{pair}}=\sum_{m<k}R_mR_k$ as the number of jointly available modality pairs. The redundancy penalty is then defined as a masked pairwise squared cosine constraint over modality-specific components:
            
            \begin{equation}
                L_{\text{red}} =
                \begin{cases}
                \beta \, \mathbb{E}\!\left[
                \dfrac{1}{N_{\mathrm{pair}}}
                \sum_{m<k} R_m R_k \cos^2(p_m,p_k)
                \right],
                & N_{\mathrm{pair}} > 0,\\[6pt]
                0,
                & N_{\mathrm{pair}} = 0.
                \end{cases}
                \label{eq:redundancy}
            \end{equation}

       Here, $\beta$ controls the strength of the penalty. When fewer than two modalities are available, $N_{\mathrm{pair}}=0$ and $L_{\text{red}}$ is set to zero. The squared cosine term penalizes both positive and negative alignment between modality-specific components, encouraging different modalities to encode distinct information rather than highly aligned representations.

\section{Experimental Evaluation}
\label{subsection:Experiments Evaluation}
    \subsection{Datasets}
    \label{subsection:datasets}
        We independently validate the proposed framework on two large-scale, multi-institutional and multimodal cohorts of Head and Neck Squamous Cell Carcinoma (HNSCC) patients, represented by the MICCAI 2025 HANCOCK \cite{dorrich2025multimodal} and the MICCAI 2025 HECKTOR \cite{andrearczyk2023overview} benchmarking environments for the tasks of predicting survival, recurrence, and HPV status. Our initial evaluation utilizes the HANCOCK dataset~\cite{dorrich2025multimodal}, a retrospective cohort of $N=763$ patients characterized by high multimodal heterogeneity and missingness. Specifically, for each patient, five potential modalities are available: primary-tumor WSI, lymph-node WSI, tissue microarrays (TMAs), unstructured pathology text reports, and structured clinical tabular data. Approximately 40\% of modality--patient pairs are missing, necessitating robust fusion strategies for the tasks of i) 2-year locoregional recurrence and ii) 5-year overall survival (OS) prediction. For independent validation and assessment of generalizability, we leverage the HECKTOR challenge dataset~\cite{andrearczyk2023overview}, comprising $N=588$ cases from nine centers. This dataset focuses on fusing molecular imaging ($^{18}$F-FDG PET), volumetric radiology (CT), and structured clinical variables. Evaluation on HECKTOR targets HPV status classification, demonstrating the framework's adaptability to diverse imaging modalities and prognostic endpoints.

    \subsection{Experimental Protocol}
    \label{subsection:experimental_protocol}
        All experiments for both HECKTOR and HANCOCK datasets were conducted using a rigorous 5-fold patient-wise cross-validation scheme and frameworks were evaluated using the Mean AUC, F1-Macro, Sensitivity, Specificity, Balanced Accuracy, and weighted F1-score. For time-to-event analyses, C-index was used as an additional ranking-based evaluation metric, using the predicted probabilities as risk scores; the model itself was trained with BCE on binarized endpoint labels rather than a censoring-aware survival loss. All frameworks were implemented in PyTorch and trained on Indiana University's HPC. Optimization was performed using the AdamW optimizer with a learning rate of $1 \times 10^{-4}$ and a weight decay of $1 \times 10^{-2}$. To stabilize the stochastic gating mechanism, the Gumbel-Sigmoid temperature $\tau$ was annealed from 1.0 to 0.1, over the first 10 epochs. The hyperparameters for the information-budget regularizer were selected by validation grid search and set to $\beta=0.01$ for the redundancy penalty and $\gamma=0.1$ for the budget penalty. Their contribution is further assessed through component ablations in Table~\ref{tab:model_elements_ablation}, where removing or isolating these terms leads to reduced performance relative to the full Multi-FRuGaL objective. Early stopping was applied with a patience of 10 epochs based on validation loss to prevent overfitting.

    \subsection{Quantitative Evaluation}
    \label{subsection:Quantitative Evaluation}

    \subsubsection{HANCOCK dataset}
    \label{subsection: HANCOCK dataset}
    
    \begin{table*}[bp]
        \centering
        \caption{Survival Comparison of multimodal fusion methods on the HANCOCK dataset.}
        \label{tab:sur_multimodal_comparison}
        \renewcommand{\arraystretch}{1.2}
        \setlength{\tabcolsep}{1pt}
    
        \begin{tabular}{@{}l l c c c c c c@{}}
        \toprule
    
        \textbf{Method} &
        \textbf{Strategy} &
        \textbf{Mean AUC} &
        \textbf{F1-Macro} &
        \textbf{Sensitivity} &
        \textbf{Specificity} &
        \textbf{Balanced Acc.} &
        \textbf{F1-weighted} \\
    
        \midrule
    
        HONeYBEE~\cite{tripathi2025honeybee} &
        Feature concat + MLP &
        \msd{0.531}{0.031} &
        \msd{0.559}{0.05} &
        \msd{0.687}{0.06} &
        \msd{0.422}{0.07} &
        \msd{0.554}{0.05} &
        \msd{0.576}{0.04} \\
    
        MulT~\cite{tsai2019multimodal} &
        Cross-Attention &
        \msd{0.543}{0.027} &
        \msd{0.567}{0.05} &
        \msd{0.692}{0.05} &
        \msd{0.444}{0.07} &
        \msd{0.567}{0.05} &
        \msd{0.597}{0.04} \\
    
        LANISTR~\cite{ebrahimi2023lanistr} &
        Generative Fusion &
        \msd{0.601}{0.021} &
        \msd{0.632}{0.04} &
        \msd{0.745}{0.05} &
        \bmsd{0.52}{0.06} &
        \msd{0.604}{0.04} &
        \msd{0.616}{0.03} \\
    
        \textbf{(Ours)} &
        \textbf{Multi-FRuGaL} &
        \bmsd{0.8496}{0.09} &
        \bmsd{0.774}{0.03} &
        \bmsd{0.883}{0.03} &
        \msd{0.51}{0.04} &
        \bmsd{0.696}{0.03} &
        \bmsd{0.801}{0.02} \\
    
        \bottomrule
        \end{tabular}
    \end{table*}

    \begin{table*}[bp]
        \centering
        \caption{Recurrence Comparison of multimodal fusion methods on the HANCOCK dataset. Multi-FRuGaL achieves the highest AUC and consistently outperforms standard concatenation, cross-attention, and generative fusion baselines.}
        \label{tab:recur_multimodal_comparison}
        \renewcommand{\arraystretch}{1.2}
        \setlength{\tabcolsep}{1pt}
    
        \begin{tabular}{@{}l l c c c c c c@{}}
        \toprule
    
        \textbf{Method} &
        \textbf{Strategy} &
        \textbf{Mean AUC} &
        \textbf{F1-Macro} &
        \textbf{Sensitivity} &
        \textbf{Specificity} &
        \textbf{Balanced Acc.} &
        \textbf{F1-weighted} \\
    
        \midrule
    
        HONeYBEE~\cite{tripathi2025honeybee} &
        Feature concat + MLP &
        \msd{0.552}{0.028} &
        \msd{0.565}{0.05} &
        \msd{0.704}{0.06} &
        \msd{0.453}{0.07} &
        \msd{0.587}{0.05} &
        \msd{0.581}{0.04} \\
    
        MulT~\cite{tsai2019multimodal} &
        Cross-Attention &
        \msd{0.575}{0.024} &
        \msd{0.584}{0.05} &
        \msd{0.723}{0.05} &
        \msd{0.471}{0.07} &
        \msd{0.598}{0.05} &
        \msd{0.606}{0.04} \\
    
        LANISTR~\cite{ebrahimi2023lanistr} &
        Generative Fusion &
        \msd{0.672}{1.9} &
        \msd{0.661}{0.04} &
        \msd{0.78}{0.05} &
        \bmsd{0.554}{0.06} &
        \bmsd{0.677}{0.04} &
        \msd{0.691}{0.03} \\
    
        \textbf{(Ours)} &
        \textbf{Multi-FRuGaL} &
        \bmsd{0.810}{0.011} &
        \bmsd{0.783}{0.03} &
        \bmsd{0.854}{0.03} &
        \msd{0.526}{0.04} &
        \msd{0.619}{0.03} &
        \bmsd{0.81}{0.02} \\
    
        \bottomrule
        \end{tabular}
    \end{table*}
 
            Across both survival and recurrence prediction tasks on the HANCOCK dataset, presented in Tables \ref{tab:sur_multimodal_comparison} and \ref{tab:recur_multimodal_comparison}, respectively, we observe highly consistent behavior across fusion strategies, with each method exhibiting similar strengths and weaknesses across endpoints. The baseline methods (i.e., HONeYBEE \cite{tripathi2025honeybee} and MulT \cite{tsai2019multimodal}) were re-implemented for HANCOCK. Since they did not account for missing data handling we incorporated a standard zero-vector imputation scheme for missing modalities. Our intention in this incorporation is to fairly compare each method by using the same modality structure with LANISTR and Multi-FRuGaL, which explicitly account for missing data. Simple feature concatenation (HONeYBEE) performs poorly in both settings, reaching only $0.531$ AUC for survival and $0.552$ for recurrence, with low specificities ($0.422$ and $0.453$), reflecting its inability to manage modality imbalance or suppress noise from high-variance WSI and text embeddings. Cross-attention (MulT) offers only marginal gains ($0.543$ and $0.575$ AUC), mirroring the survival results: without redundancy control or sample-wise gating, it distributes attention uniformly across heterogeneous and unaligned modalities, diluting informative signals. LANISTR consistently emerges as the strongest baseline, achieving $0.601$ AUC (survival) and $0.672$ AUC (recurrence) with the highest baseline specificity ($0.52$ and $0.554$), indicating that latent-space harmonization stabilizes cross-modality variance but still plateaus due to over-imputation and a lack of redundancy handling. A similar trend is observed in the survival ranking performance (Table~\ref{tab:C_index_multimodal_comparison_hancock}), where the strongest baseline achieves C-indices of $0.6107$ for OS, $0.6915$ for recurrence-free survival (RFS), and $0.6815$ for progression-free survival (PFS). This behavior is consistent with the limitations of existing multimodal strategies discussed in the related work section \ref{section:Related Work} earlier: generative fusion relies on imputed representations that may deviate from underlying biological signals, while attention-based and concatenation methods lack explicit mechanisms to regulate modality reliability or suppress redundant information. In contrast, Multi-FRuGaL maintains a clear margin across both tasks, achieving \textbf{$0.8496$ AUC} for survival and \textbf{$0.8102$} for recurrence, with the highest sensitivities ($0.883$ and $0.854$) and strongest F1-weighted scores ($0.801$ and $0.810$). This advantage is further supported by survival ranking performance, where Multi-FRuGaL achieves the highest C-indices reaching \textbf{$0.6814 \pm 0.015$} for OS, \textbf{$0.7421 \pm 0.013$} for RFS, and \textbf{$0.7143 \pm 0.015$} for PFS. These consistent improvements demonstrate the efficacy of Multi-FRuGaL’s signal decomposition with a modality-level gating mechanism, which suppresses noisy modalities while explicitly penalizing redundancy. By enforcing sparse, complementary information usage, our framework generalizes robustly across both long-term survival and short-horizon recurrence prediction.
            
            \begin{table*}[t]
                \centering
                \caption{Comparison of multimodal fusion methods on the HECKTOR dataset for HPV status prediction. Multi-FRuGaL outperforms feature concatenation, cross-attention, and CNN+MLP baselines.}
                \label{tab:hecktor_comparison}
                \renewcommand{\arraystretch}{1.0}
                \setlength{\tabcolsep}{1pt}
                \begin{tabular}{@{}l l c c c c c c@{}}
                \toprule
                \textbf{Method} &
                \textbf{Strategy} &
                \textbf{Mean AUC} &
                \textbf{F1-Macro} &
                \textbf{Sensitivity} &
                \textbf{Specificity} &
                \textbf{Balanced Acc.} &
                \textbf{F1-weighted} \\
                \midrule
            
                EfficientNet~\cite{tan2019efficientnet} &
                Feature concat + MLP &
                \msd{0.581}{0.25} &
                \msd{0.556}{0.11} &
                \msd{0.993}{0.03} &
                \msd{0.122}{0.16} &
                \msd{0.551}{0.07} &
                \msd{0.871}{0.03} \\
            
                HONeYBEE~\cite{tripathi2025honeybee} &
                Feature concat + MLP &
                \msd{0.734}{0.31} &
                \msd{0.616}{0.10} &
                \msd{0.981}{0.03} &
                \msd{0.183}{0.14} &
                \msd{0.584}{0.08} &
                \msd{0.885}{0.03} \\
            
                DenseNet121~\cite{huang2017densenet} &
                Feature concat + MLP &
                \msd{0.777}{0.22} &
                \msd{0.646}{0.13} &
                \bmsd{0.994}{0.04} &
                \msd{0.223}{0.19} &
                \msd{0.617}{0.10} &
                \msd{0.891}{0.04} \\
            
                SENet~\cite{hu2018senet} &
                Feature concat + MLP &
                \msd{0.792}{0.04} &
                \msd{0.584}{0.11} &
                \msd{0.993}{0.01} &
                \msd{0.151}{0.15} &
                \msd{0.575}{0.07} &
                \msd{0.871}{0.03} \\
            
                ViT~\cite{dosovitskiy2021vit} &
                Feature concat + MLP &
                \msd{0.852}{0.05} &
                \msd{0.521}{0.11} &
                \msd{1.00}{0.00} &
                \msd{0.076}{0.15} &
                \msd{0.534}{0.07} &
                \msd{0.867}{0.03} \\
            
                LANISTR~\cite{ebrahimi2023lanistr} &
                Generative Fusion &
                \msd{0.854}{0.086} &
                \msd{0.661}{0.09} &
                \msd{0.953}{0.03} &
                \msd{0.324}{0.11} &
                \msd{0.642}{0.06} &
                \msd{0.891}{0.03} \\
            
                Less-is-More~\cite{cailess} &
                CNN + MLP &
                \msd{0.861}{0.21} &
                \msd{0.736}{0.07} &
                \msd{0.971}{0.02} &
                \msd{0.364}{0.10} &
                \msd{0.675}{0.05} &
                \msd{0.913}{0.02} \\
            
                MulT~\cite{tsai2019multimodal} &
                Cross-Attention &
                \msd{0.904}{0.09} &
                \msd{0.691}{0.08} &
                \msd{0.963}{0.02} &
                \msd{0.415}{0.12} &
                \msd{0.684}{0.06} &
                \msd{0.906}{0.02} \\
            
                ResNet18~\cite{he2016resnet} &
                Feature concat + MLP &
                \msd{0.946}{0.03} &
                \msd{0.774}{0.09} &
                \msd{0.993}{0.00} &
                \msd{0.432}{0.15} &
                \msd{0.718}{0.07} &
                \msd{0.936}{0.02} \\
            
                \textbf{Ours} &
                \textbf{Multi-FRuGaL} &
                \bmsd{0.975}{0.06} &
                \bmsd{0.841}{0.05} &
                \msd{0.992}{0.01} &
                \bmsd{0.512}{0.10} &
                \bmsd{0.753}{0.04} &
                \bmsd{0.941}{0.02} \\
            
                \bottomrule
                \end{tabular}
            \end{table*}

            \begin{table*}[bp]
                \centering
                \caption{Concordance Index for HANCOCK}
                \label{tab:C_index_multimodal_comparison_hancock}
                \renewcommand{\arraystretch}{1.2}
                \setlength{\tabcolsep}{1pt}
                \begin{tabular}{@{}l c c c@{}}
                \toprule
                \textbf{Method} &
                \textbf{OS} &
                \textbf{Recurrence Free Survival} &
                \textbf{Progression Free Survival} \\
                \midrule
            
                HONeYBEE~\cite{tripathi2025honeybee} &
                \msd{0.5401}{0.024} &
                \msd{0.6372}{0.033} &
                \msd{0.5941}{0.024} \\
            
                MulT~\cite{tsai2019multimodal} &
                \msd{0.5926}{0.042} &
                \msd{0.6603}{0.012} &
                \msd{0.6766}{0.042} \\
            
                LANISTR~\cite{ebrahimi2023lanistr} &
                \msd{0.6107}{0.021} &
                \msd{0.6615}{0.021} &
                \msd{0.6515}{0.021} \\
            
                \textbf{(Ours)} &
                \bmsd{0.6814}{0.015} &
                \bmsd{0.7421}{0.013} &
                \bmsd{0.7143}{0.015} \\
            
                \bottomrule
                \end{tabular}
            \end{table*}

            \begin{table*}[bp]
                \centering
                \caption{Concordance Index for HECKTOR.}
                \label{tab:C_index_multimodal_comparison_HECKTOR}
                \renewcommand{\arraystretch}{1.2}
                \setlength{\tabcolsep}{1pt}
                \begin{tabular}{@{}l c@{}}
                \toprule
                \textbf{Method} &
                \textbf{Recurrence Free Survival} \\
                \midrule
            
                HONeYBEE~\cite{tripathi2025honeybee} &
                \msd{0.5842}{0.044} \\
            
                MulT~\cite{tsai2019multimodal} &
                \msd{0.6473}{0.051} \\
            
                LANISTR~\cite{ebrahimi2023lanistr} &
                \msd{0.6323}{0.032} \\
            
                \textbf{(Ours)} &
                \bmsd{0.7203}{0.022} \\
            
                \bottomrule
                \end{tabular}
            \end{table*}
                        
        \subsubsection{HECKTOR Dataset}
        \label{subsection:HECKTOR}
            The HPV prediction results on the HECKTOR dataset (Table \ref{tab:hecktor_comparison}) reveal a clear hierarchy among fusion strategies, with trends that echo but also amplify the observations from HANCOCK. The baseline methods including EfficientNet \cite{tan2019efficientnet}, HONeYBEE \cite{tripathi2025honeybee}, DenseNet121 \cite{huang2017densenet}, SENet \cite{hu2018senet}, ViT \cite{dosovitskiy2021vit}, Less-is-More \cite{cailess}, MulT \cite{tsai2019multimodal}, and ResNet18 \cite{he2016resnet} rely on fixed feature concatenation followed by an MLP, and while deeper backbones improve representation quality, they remain fundamentally limited by their inability to adaptively regulate modality contributions. HONeYBEE and traditional CNN-MLP pipelines (such as DenseNet121 \cite{huang2017densenet} and SENet \cite{hu2018senet}) achieve only moderate performance (AUCs of $0.734$, $0.777$, and $0.792$), with extremely high sensitivity ($0.981$--$0.994$) but very low specificity ($0.122$--$0.223$), indicating a strong bias toward predicting HPV-positive cases. Larger or more expressive backbones (such as ViT \cite{dosovitskiy2021vit} and EfficientNet \cite{tan2019efficientnet} do not alleviate this imbalance; EfficientNet in particular collapses to an AUC of $0.581$. Among deterministic baselines, ResNet18 \cite{he2016resnet} stands out with a much stronger AUC of $0.946$, while still suffering from low specificity ($0.432$). Cross-attention (MulT) \cite{tsai2019multimodal} performs competitively ($0.904$ AUC), again confirming its ability to capture modality interactions, yet without mechanisms for sparsity or redundancy control, its specificity remains modest. LANISTR \cite{ebrahimi2023lanistr} and Less-is-More \cite{cailess} provide further improvements through latent harmonization and task-specific CNN features (AUCs $0.854$ and $0.861$), but both methods still underperform on specificity and balanced accuracy. A similar pattern is observed in survival ranking performance (Table~\ref{tab:C_index_multimodal_comparison_HECKTOR}), where the strongest baseline reaches a C-index of $0.6473$ for RFS. This pattern reflects the limitations outlined in the related work section \ref{section:Related Work} earlier: fixed fusion strategies and latent harmonization improve average performance but lack explicit mechanisms to regulate modality reliability or suppress redundant radiological signals, leading to imbalanced decision boundaries under heterogeneous clinical presentations. In contrast, our Multi-FRuGaL framework achieves the highest overall performance with \textbf{0.975 AUC}, \textbf{0.841 F1-Macro}, and the strongest balanced accuracy ($0.753$). This advantage is further supported by survival ranking performance, where Multi-FRuGaL achieves the highest RFS C-index, reaching \textbf{$0.7203 \pm 0.022$}. These gains reflect the benefits of signal decomposition, modality-level gating, and information-budgeted loss objectives: the framework can separate modality-shared and modality-specific signals, selectively amplify PET or CT features when clinically informative while suppressing redundancy and noise, yielding both high sensitivity ($0.992$) and a substantial improvement in specificity ($0.512$) relative to all baselines. The results highlight that Multi-FRuGaL not only outperforms concatenation and cross-attention methods but also brings a more balanced decision boundary for HPV classification.

        \subsubsection{Ablation studies}
        \label{subsection:Ablation studies}
    
        \begin{table*}[t]
            \centering
            \caption{Modality ablation study on HANCOCK (recurrence and survival) and HECKTOR (HPV status). For HANCOCK, primary-tumor WSI and lymph-node WSI are grouped as a single WSI stream for modality-level ablation.}
            \label{tab:modality_ablation}
            \renewcommand{\arraystretch}{1.0}
            \setlength{\tabcolsep}{1pt}
        
            \begin{tabular}{@{}l l c c c c@{}}
            \toprule
        
            & \multicolumn{3}{c}{\textbf{HANCOCK}} &
            \multicolumn{2}{c}{\textbf{HECKTOR}} \\
        
            \cmidrule(lr){2-4} \cmidrule(lr){5-6}
        
            \textbf{Category} &
            \textbf{Modality} &
            \textbf{AUC Recurrence} &
            \textbf{AUC Survival} &
            \textbf{Modality} &
            \textbf{HPV AUC} \\
        
            \midrule
        
            {\textbf{Unimodal}}
            & WSI (Prim + Lymph)
            & \msd{0.6303}{0.021}
            & \msd{0.6153}{0.024}
            & CT
            & \msd{0.7557}{0.041}\\
        
            & Text
            & \msd{0.6042}{0.018}
            & \msd{0.6016}{0.019}
            & PET
            & \msd{0.6803}{0.052}\\
        
            & TMA
            & \msd{0.5911}{0.020}
            & \msd{0.6662}{0.022}
            & Clinical
            & \msd{0.9734}{0.018}\\
        
            & Clinical Tables
            & \msd{0.8002}{0.016}
            & \msd{0.8209}{0.015}
            & 
            & \\
        
            \midrule
        
            {\textbf{Bimodal}}
            & WSI + Text
            & \msd{0.6728}{0.019}
            & \msd{0.6467}{0.018}
            & PET+Clinical
            & \msd{0.9075}{0.029}\\
        
            & WSI + TMA
            & \msd{0.6541}{0.020}
            & \msd{0.7052}{0.019}
            & CT+PET
            & \msd{0.784}{0.037}\\
        
            & WSI + Clinical Tables
            & \msd{0.7894}{0.014}
            & \msd{0.8117}{0.013}
            & CT+Clinical
            & \bmsd{0.9765}{0.014}\\
        
            & Text + TMA
            & \msd{0.6125}{0.018}
            & \msd{0.6764}{0.020}
            &
            & \\
        
            & Text + Clinical Tables
            & \msd{0.7974}{0.017}
            & \msd{0.8152}{0.015}
            &
            & \\
        
            & TMA + Clinical Tables
            & \msd{0.8037}{0.014}
            & \msd{0.8251}{0.013}
            &
            & \\
        
            \midrule
        
            {\textbf{Trimodal}}
            & WSI + Text + TMA
            & \msd{0.6832}{0.018}
            & \msd{0.7113}{0.018}
            & CT+PET+Clinical
            & \msd{0.9757}{0.013}\\
        
            & WSI + Text + Clinical Tables
            & \msd{0.7829}{0.015}
            & \msd{0.8038}{0.014}
            &
            & \\
        
            & WSI + TMA + Clinical Tables
            & \msd{0.7991}{0.013}
            & \msd{0.8314}{0.012}
            &
            & \\
        
            & Text + TMA + Clinical Tables
            & \msd{0.7967}{0.014}
            & \msd{0.8163}{0.013}
            &
            & \\
        
            \midrule
        
            {\textbf{Quadmodal}}
            & Tabular + Text + WSI + TMA
            & \bmsd{0.8102}{0.012}
            & \bmsd{0.8496}{0.011}
            &
            & \\
        
            \bottomrule
            \end{tabular}
        \end{table*}

\begin{table*}[t]
    \centering
    \caption{Ablation study of framework components on HANCOCK (Recurrence, Survival) and HECKTOR (HPV Status).}
    \label{tab:model_elements_ablation}
    \renewcommand{\arraystretch}{1.2}
    \setlength{\tabcolsep}{1pt}

    \begin{tabular}{@{}l c c c@{}}
    \toprule

    & \multicolumn{2}{c}{\textbf{HANCOCK}} &
    \textbf{HECKTOR (40\% missing)} \\

    \cmidrule(lr){2-3}
    \cmidrule(lr){4-4}

    Base Model + \textbf{Component(s)} &
    \textbf{AUC Recurrence} &
    \textbf{AUC Survival} &
    \textbf{HPV Status AUC} \\

    \midrule

    GateNet
    & \msd{0.7521}{0.015}
    & \msd{0.7945}{0.014}
    & \msd{0.8812}{0.031} \\

    SD Layer + GateNet
    & \msd{0.7768}{0.014}
    & \msd{0.8106}{0.013}
    & \msd{0.9187}{0.028} \\

    GateNet + $L_{\mathrm{budget}}$
    & \msd{0.7545}{0.014}
    & \msd{0.8023}{0.013}
    & \msd{0.9240}{0.029} \\

    GateNet + $L_{\mathrm{red}}$
    & \msd{0.7692}{0.013}
    & \msd{0.7880}{0.012}
    & \msd{0.9085}{0.028} \\

    SD Layer + GateNet + $L_{\mathrm{budget}}$
    & \msd{0.7892}{0.013}
    & \msd{0.8215}{0.012}
    & \msd{0.9316}{0.027} \\

    SD Layer + GateNet + $L_{\mathrm{red}}$
    & \msd{0.7968}{0.012}
    & \msd{0.8317}{0.011}
    & \msd{0.9408}{0.026} \\\midrule

    \textbf{\shortstack[l]{(Multi-FRuGaL)\\SD Layer + GateNet + $L_{\mathrm{IBR}}$ + $L_{\mathrm{dec}}$}}
    & \bmsd{0.8102}{0.012}
    & \bmsd{0.8496}{0.011}
    & \bmsd{0.9513}{0.026} \\

    \bottomrule
    \end{tabular}
\end{table*}

\paragraph{\textbf{Modality ablation study}}
For HANCOCK, primary-tumor WSI and lymph-node WSI were processed as separate WSI-derived inputs during feature extraction and fusion, but were grouped as a single WSI stream in the modality-level ablation analysis to simplify comparison across modality combinations. To gain a better understanding of the contribution of each input stream, we conducted a comprehensive modality ablation study across both HANCOCK (recurrence and survival) and HECKTOR (HPV status), evaluating Multi-FRuGaL under unimodal, bimodal, trimodal, and full-modality settings (Table~\ref{tab:modality_ablation}). On HANCOCK, unimodal performance varies substantially across modalities, with clinical tabular data emerging as the strongest single source ($0.800$--$0.821$ AUC). This aligns with clinical intuition, as these variables (e.g., TNM staging, surgical margins) represent high-level semantic distillations of diagnostic findings and therefore encode strong prognostic priors. Conversely, raw modalities like WSI, TMA, and text, provide weaker but complementary signals, capturing granular morphological and biological heterogeneity not fully reflected in the tabular summaries. As modalities are combined, performance rises steadily, reflecting the value of cross-modal complementarities: bimodal clinical+WSI combinations reach up to $0.825$ survival AUC, and trimodal sets that include clinical features consistently outperform purely image-based subsets. A similar pattern appears in HECKTOR, where clinical features alone already achieve a strong HPV classification ($0.973$ AUC), PET+clinical reaches $0.907$, and the full CT+PET+clinical fusion remains highly competitive ($0.976$). Similarly with HANCOCK, clinical data emerge as the strongest single modality, as they are distillations of diagnostic findings, i.e., incorporating manually observed priors. Collectively, these results demonstrate that no single modality is sufficient, but specific combinations, especially those involving clinical data produce the most discriminative representations. The consistent monotonic improvement as modalities are added further supports the design of Multi-FRuGaL, which is built to exploit complementary signals while suppressing redundancy.

            \paragraph{\textbf{Component ablation studies}}
                To further quantitatively evaluate the contribution of each architectural component in our proposed Multi-FRuGaL, we performed an elements ablation study by selectively disabling GateNet, the redundancy penalty ($L_{\mathrm{red}}$), and the information budget ($L_{\mathrm{budget}}$), evaluating the resulting frameworks on both HANCOCK outcomes and HECKTOR HPV classification (Table~\ref{tab:model_elements_ablation}). Removing GateNet leads to a substantial drop in performance across all tasks (e.g., recurrence AUC falls to $0.752$ and HPV AUC to $0.881$), confirming the central role of sample-wise modality gating in suppressing noisy or uninformative inputs. Eliminating the redundancy penalty has an even larger impact ($0.684$ recurrence AUC; $0.824$ HPV AUC), reflecting the importance of discouraging overlapping signals, particularly between highly correlated image modalities, such as WSI and TMA. Combining GateNet with the budget loss improves performance, but still lags behind the full framework, indicating that sparsity alone is insufficient without redundancy management. The full Multi-FRuGaL configuration, which integrates SD layer, GateNet, $L_{\mathrm{dec}}$, and $L_{\mathrm{IBR}}$, achieves the strongest results across all datasets and tasks (e.g., $0.8102$ recurrence AUC, $0.8496$ survival AUC, and $0.9513$ HPV AUC). These findings demonstrate that Multi-FRuGaL’s components are complementary: gating provides selective uptake of informative modalities, the budget enforces efficient use of information, and the redundancy penalty prevents representational overlap, collectively producing the most robust multimodal fusion.

\begin{figure*}[t]
                \centering
                 \includegraphics[width=0.6\textwidth]{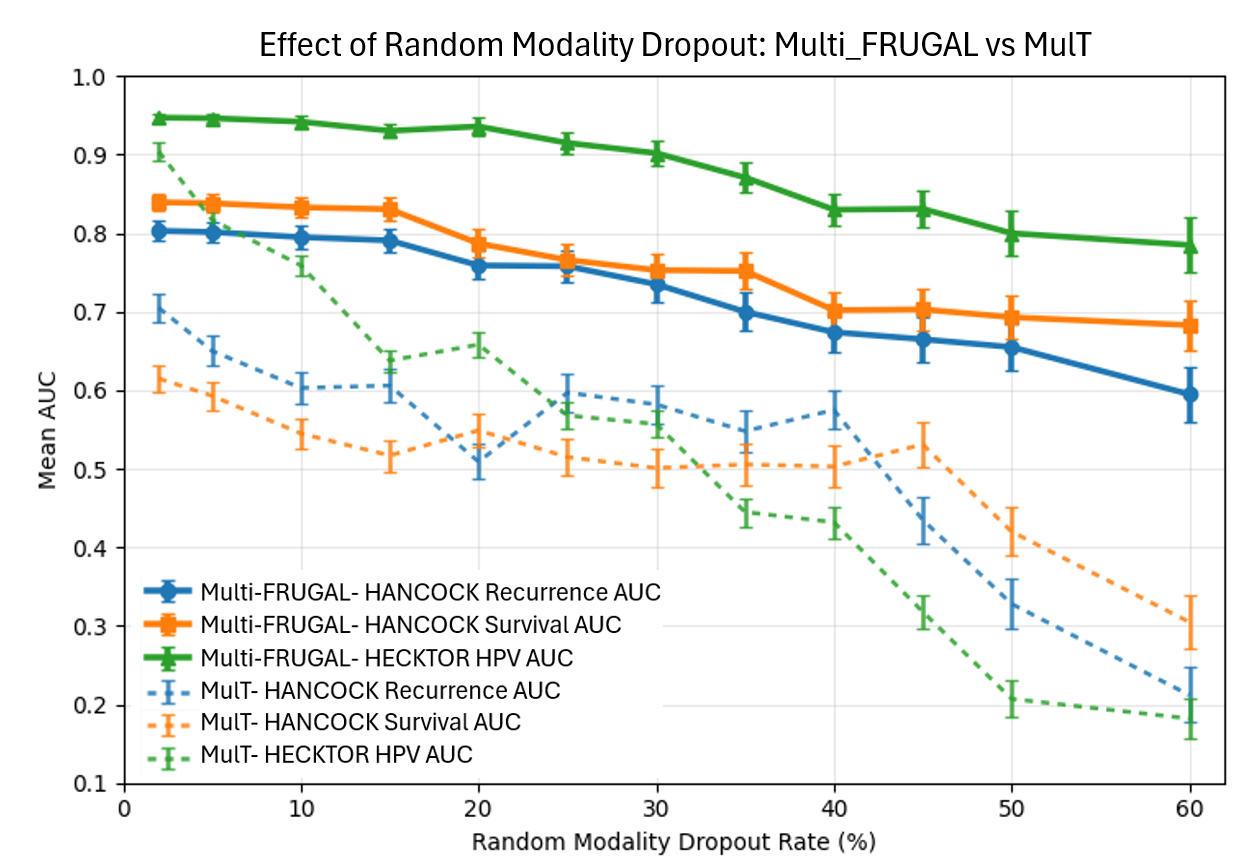}
                \caption{Performance under random modality dropout across HANCOCK recurrence, HANCOCK survival, and HECKTOR HPV prediction.}
                \label{fig:robustness_plot}
            \end{figure*}

            \paragraph{\textbf{Modality missingness analysis}}
                To assess robustness under increasing data missingness, we conducted a random modality dropout study, where a controlled fraction of available modalities was randomly removed at inference time (Fig.~\ref{fig:robustness_plot}). We compare Multi-FRuGaL against MulT \cite{tsai2019multimodal}, a representative cross-attention–based fusion method that lacks explicit mechanisms for handling missing modalities. Across all tasks, Multi-FRuGaL exhibits a smooth and gradual degradation in performance as missingness increases, whereas MulT degrades substantially faster. On HANCOCK's recurrence task, Multi-FRuGaL AUC decreases from $0.803$ at 2\% dropout to $0.595$ at 60\%, while at the survival task the AUC drops from $0.840$ to $0.683$. In contrast, MulT suffers an early and sharp decline, falling below $0.55$ AUC by 40\% dropout for both outcomes. A similar trend is observed on HECKTOR's HPV task, where Multi-FRuGaL maintains AUC above $0.90$ up to 30\% dropout and remains competitive ($0.785$) even at 60\%, whereas MulT rapidly deteriorates from $0.904$ at low dropout to below $0.45$ under severe missingness. This divergence highlights the limitations of cross-attention when modalities are absent, as attention weights become ill-defined without explicit gating. By contrast, Multi-FRuGaL’s modality-level gating mechanism formulation dynamically suppresses missing or low-value inputs and reallocates capacity to informative modalities, enabling stable and robust fusion even under aggressive modality dropout.
                
    \subsection{Qualitative Analysis}
        \subsubsection{HANCOCK Dataset}
\begin{figure*}[t]
    \centering

    \newcommand{\kmw}{0.30\textwidth}
    \newcommand{\kmh}{0.17\textheight}

    \newcommand{\kmplot}[1]{%
        \adjustbox{valign=m}{%
            \includegraphics[width=\kmw,height=\kmh,keepaspectratio]{#1}%
        }%
    }

    \newcommand{\methodname}[1]{%
        \adjustbox{valign=m}{%
            \rotatebox[origin=c]{90}{\textbf{#1}}%
        }%
    }

    \setlength{\tabcolsep}{2pt}
    \renewcommand{\arraystretch}{1.05}

    \begin{tabular}{c c c c}
        \toprule
        \textbf{Model} & \textbf{OS} & \textbf{RFS} & \textbf{PFS} \\
        \midrule

        \methodname{HONeYBEE} &
        \kmplot{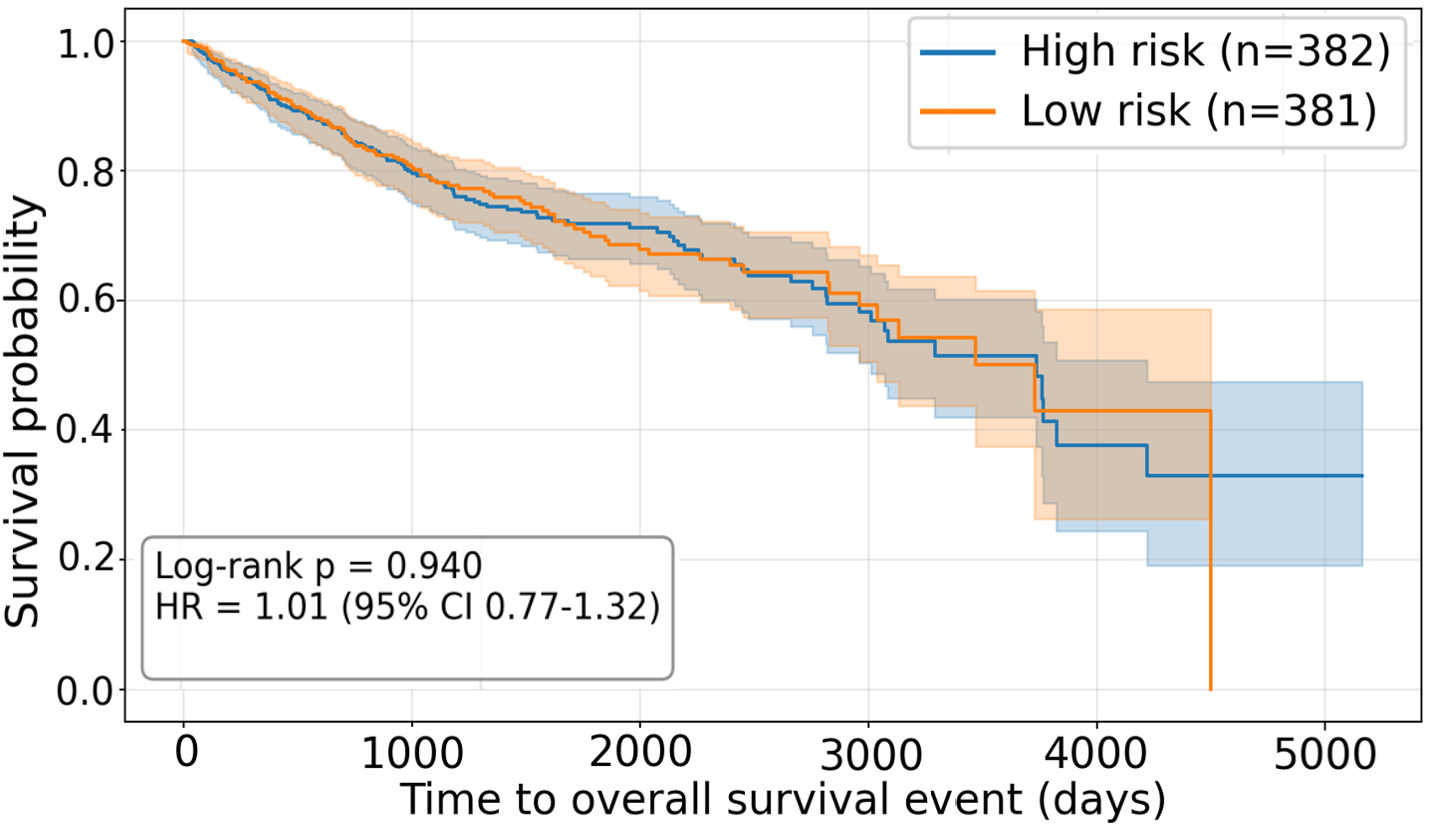} &
        \kmplot{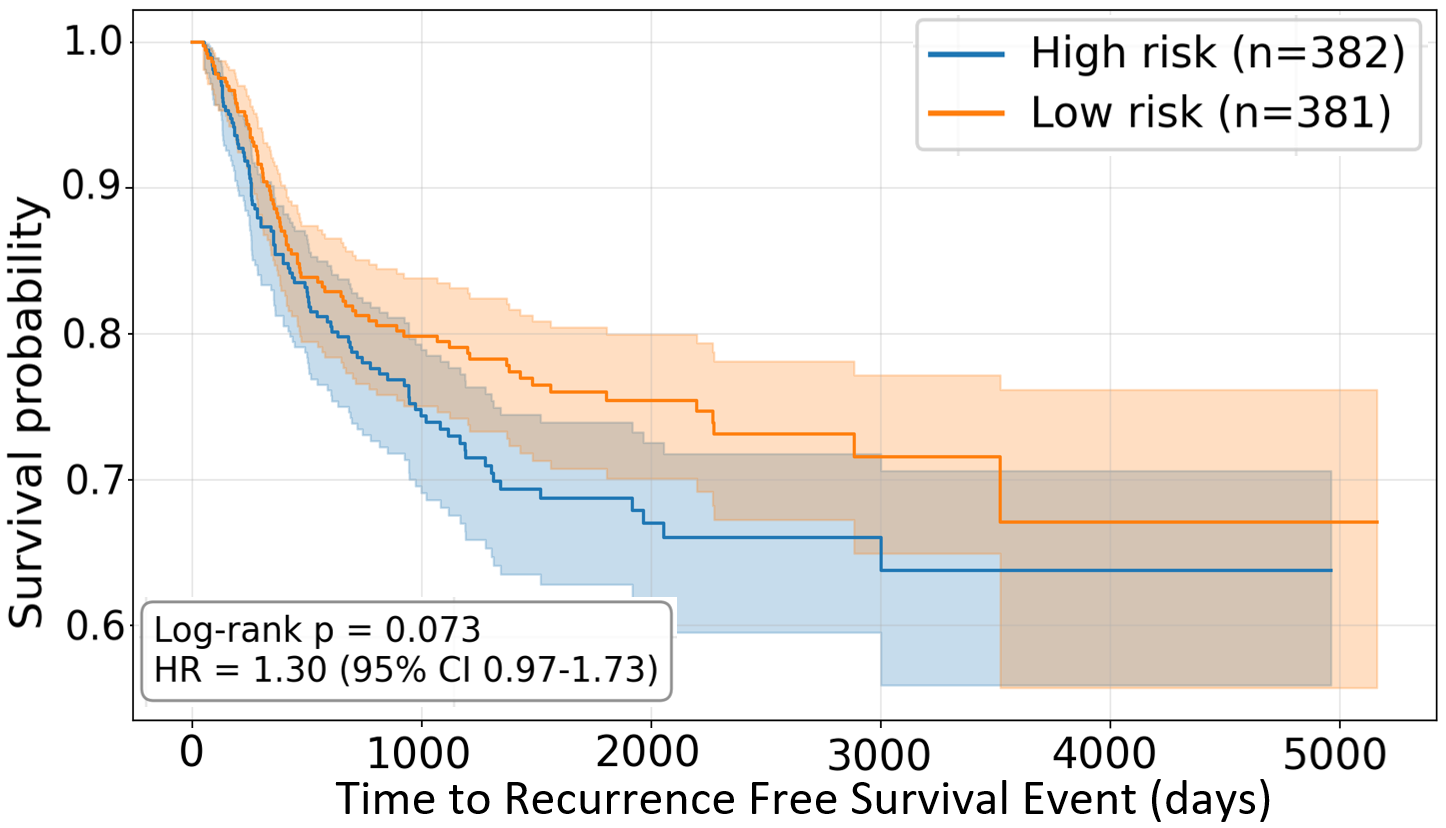} &
        \kmplot{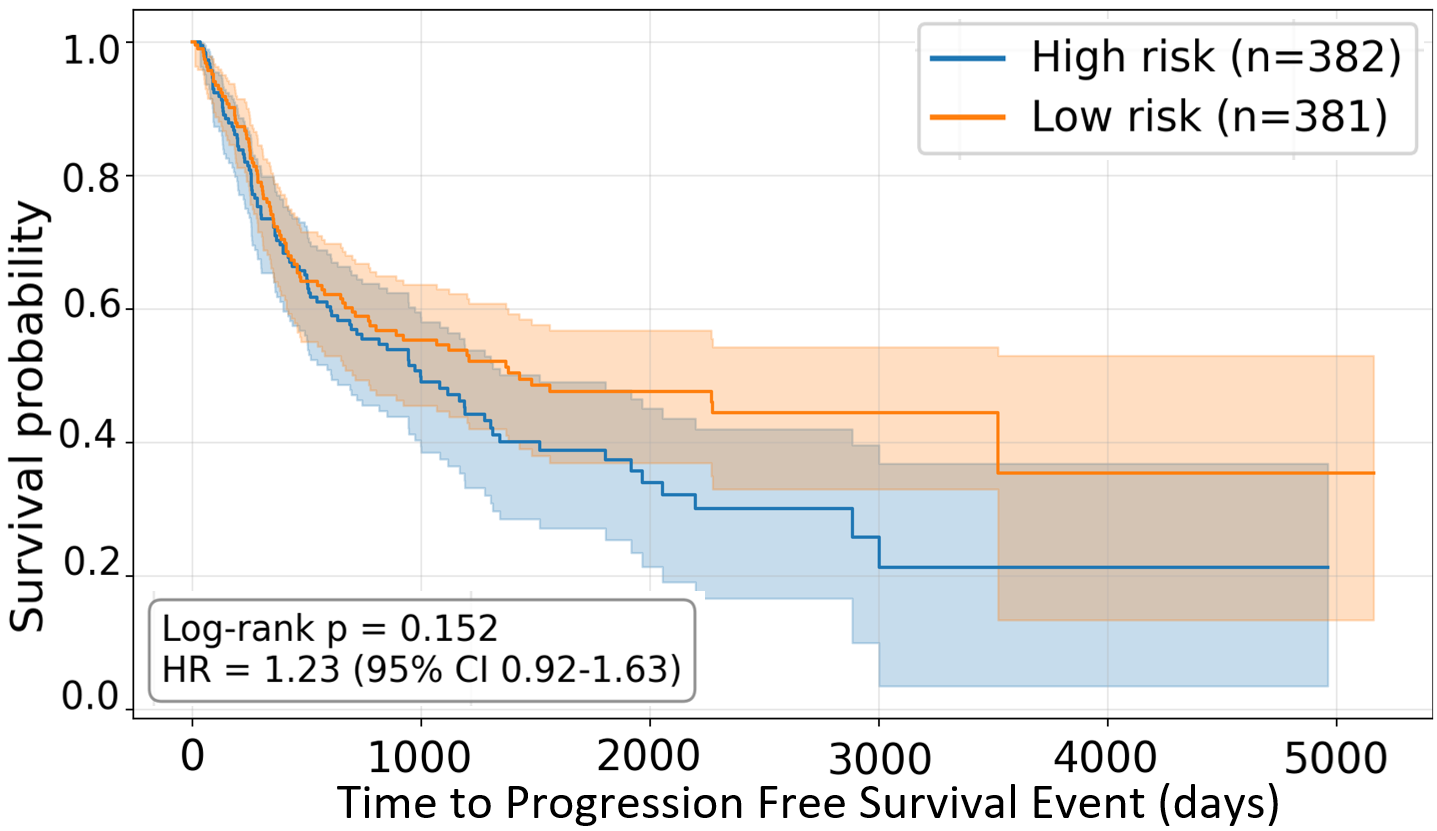} \\

        \methodname{MulT} &
        \kmplot{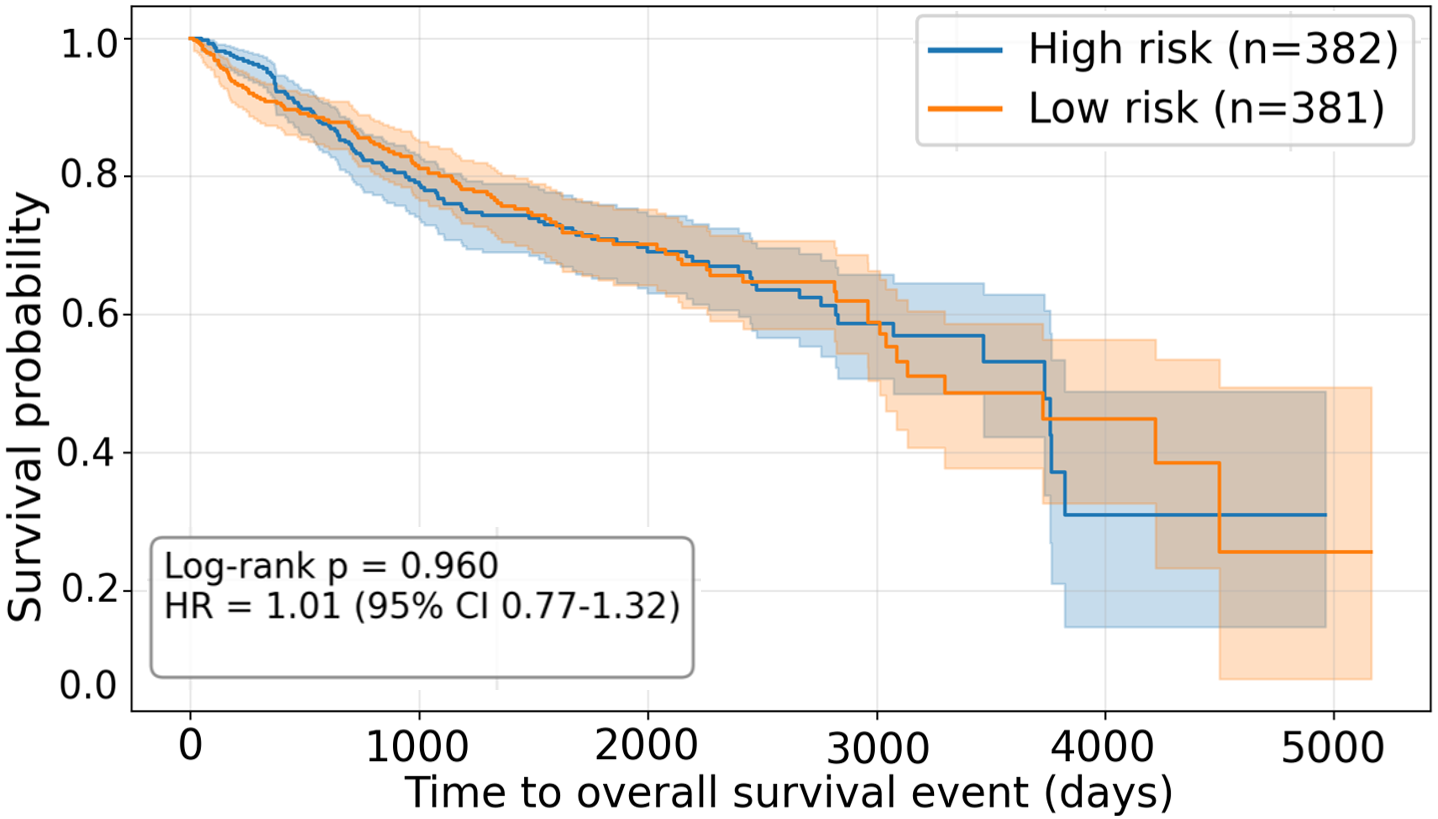} &
        \kmplot{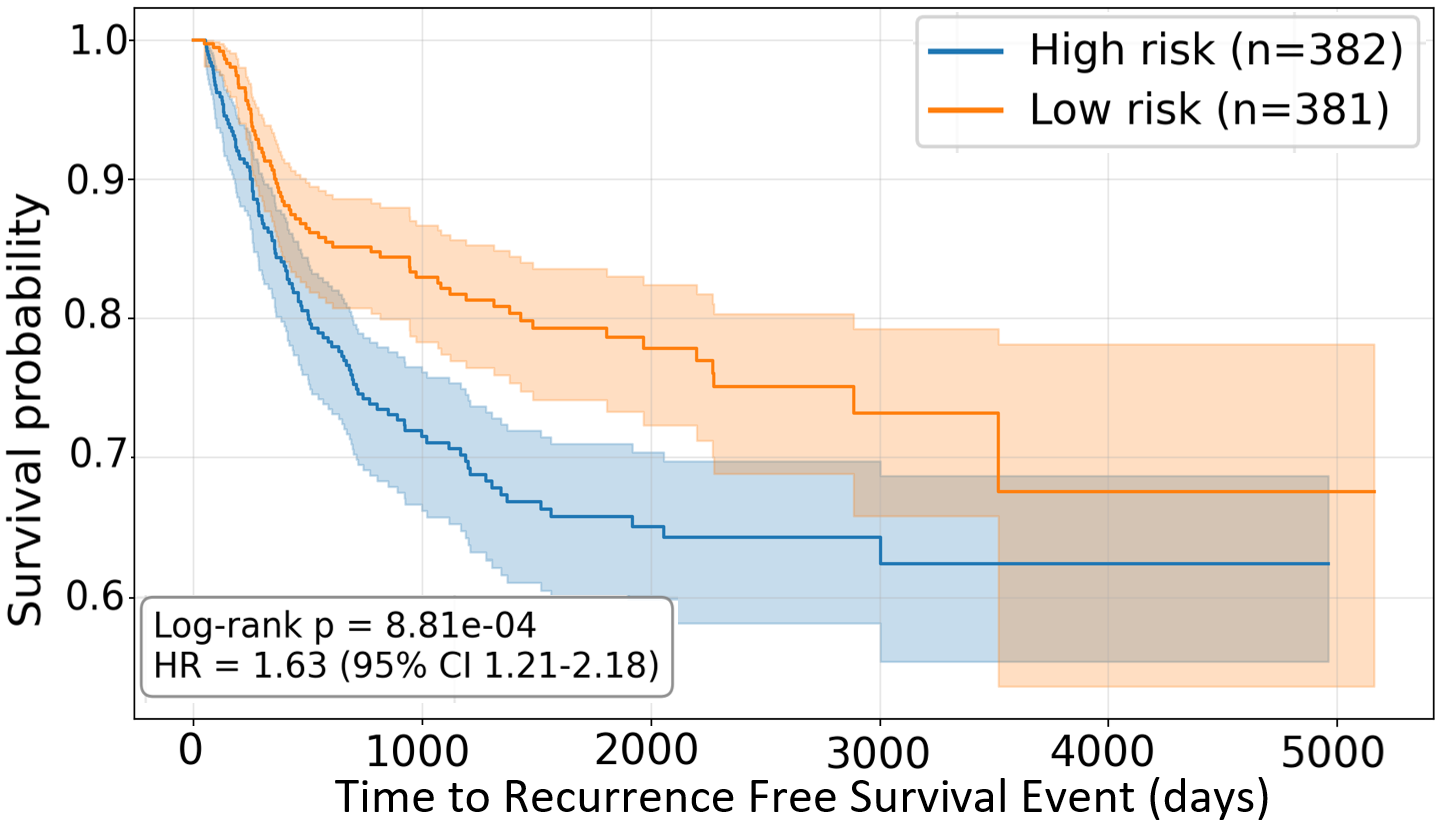} &
        \kmplot{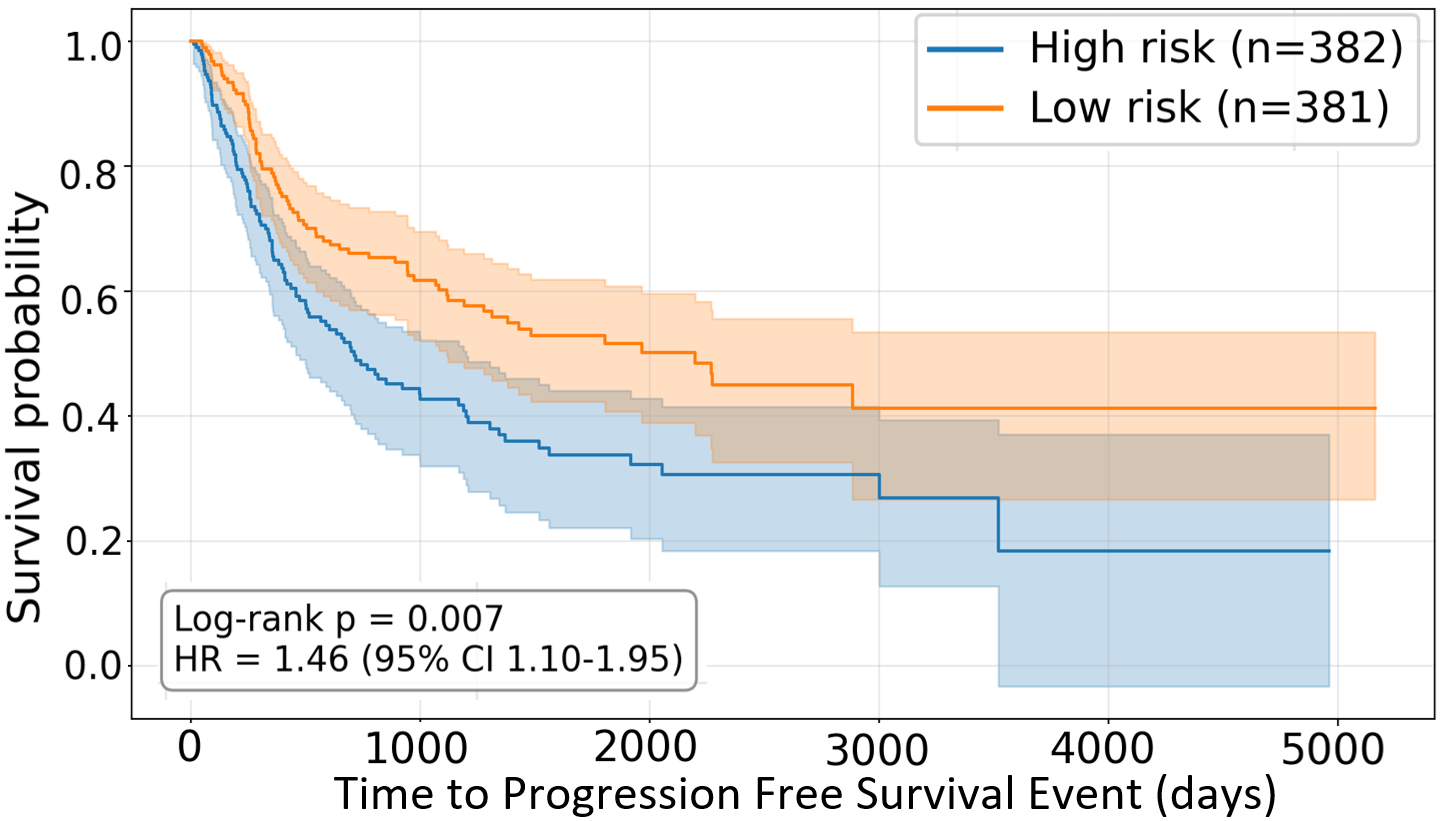} \\

        \methodname{LANISTR} &
        \kmplot{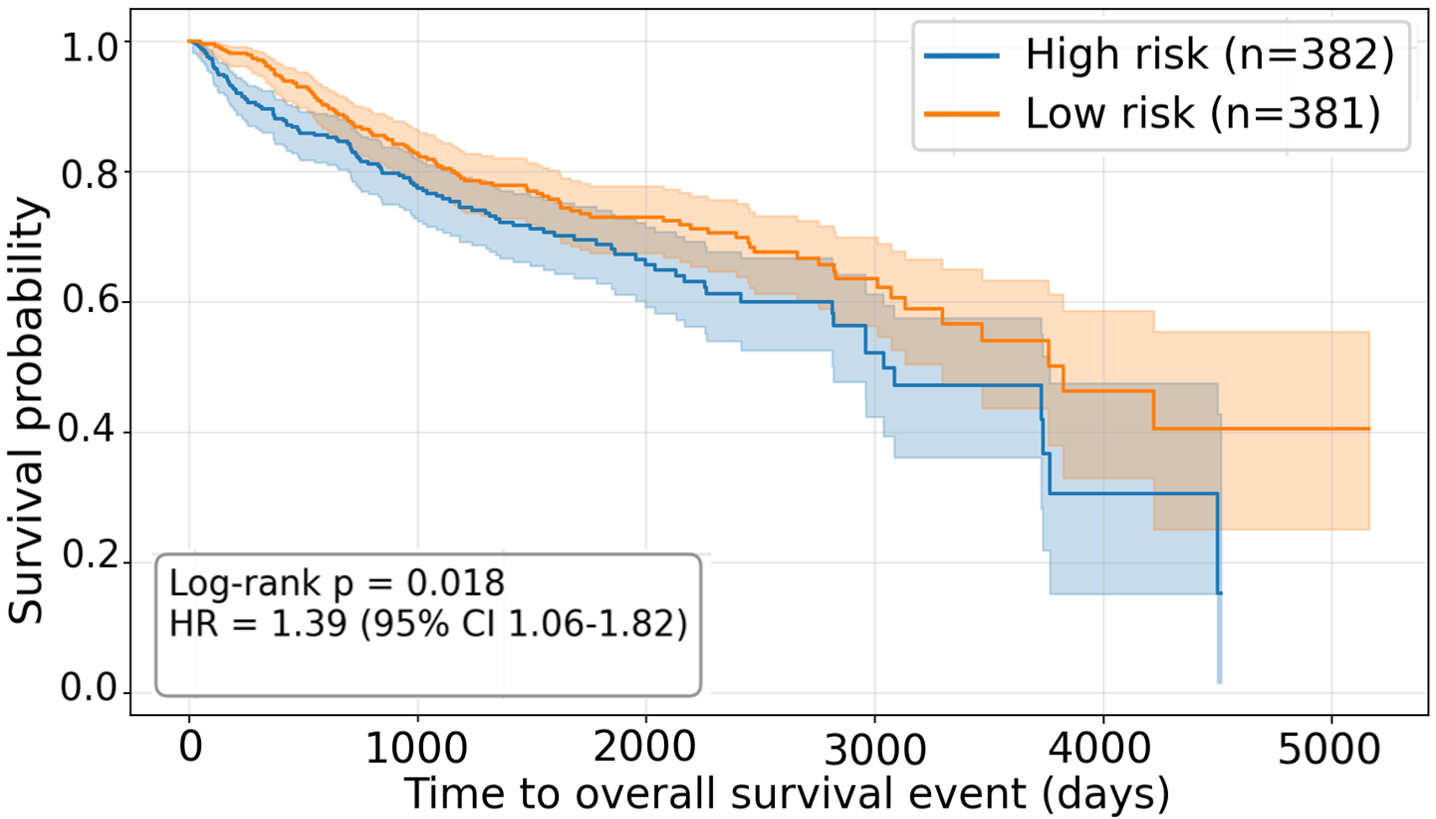} &
        \kmplot{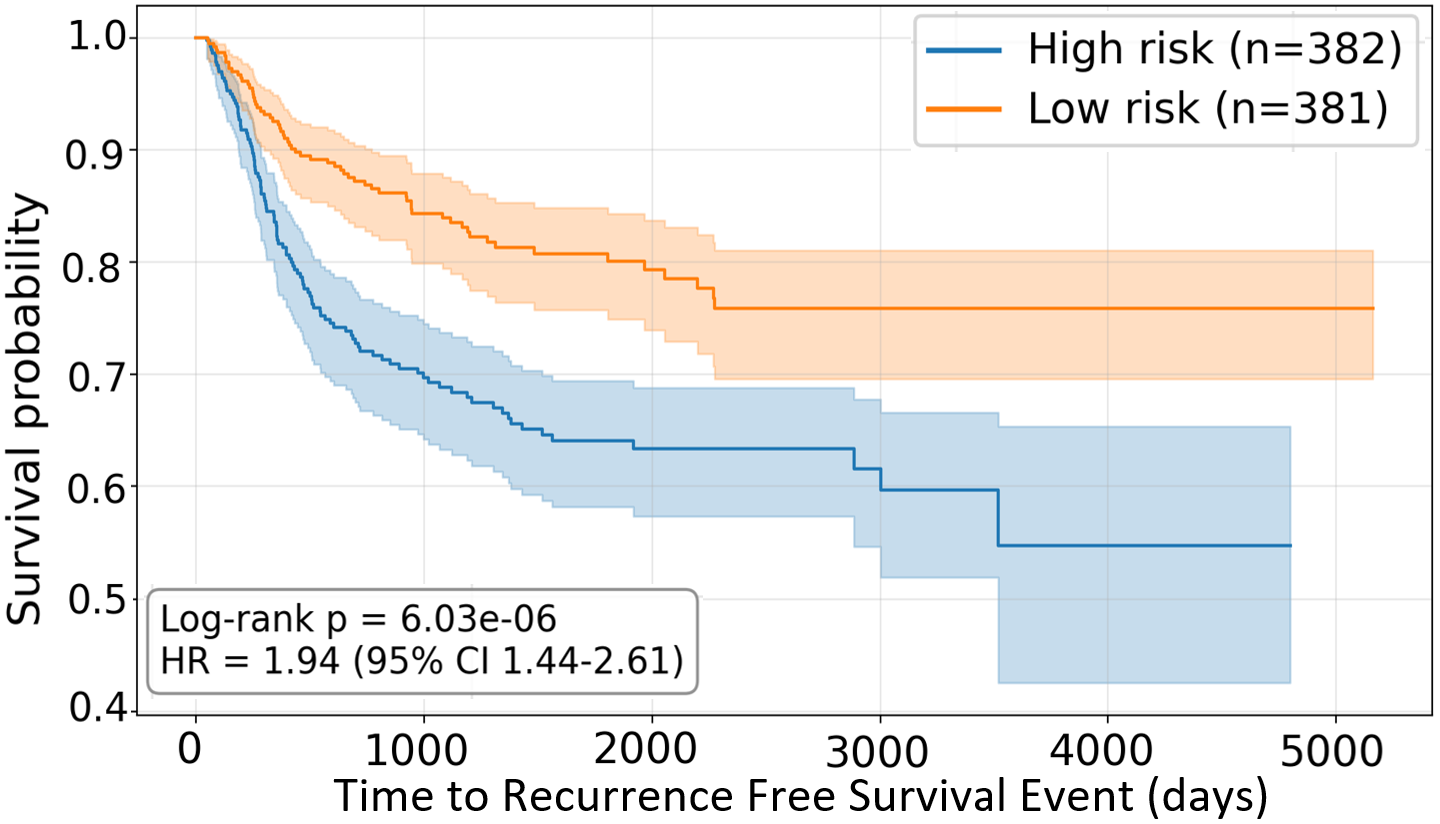} &
        \kmplot{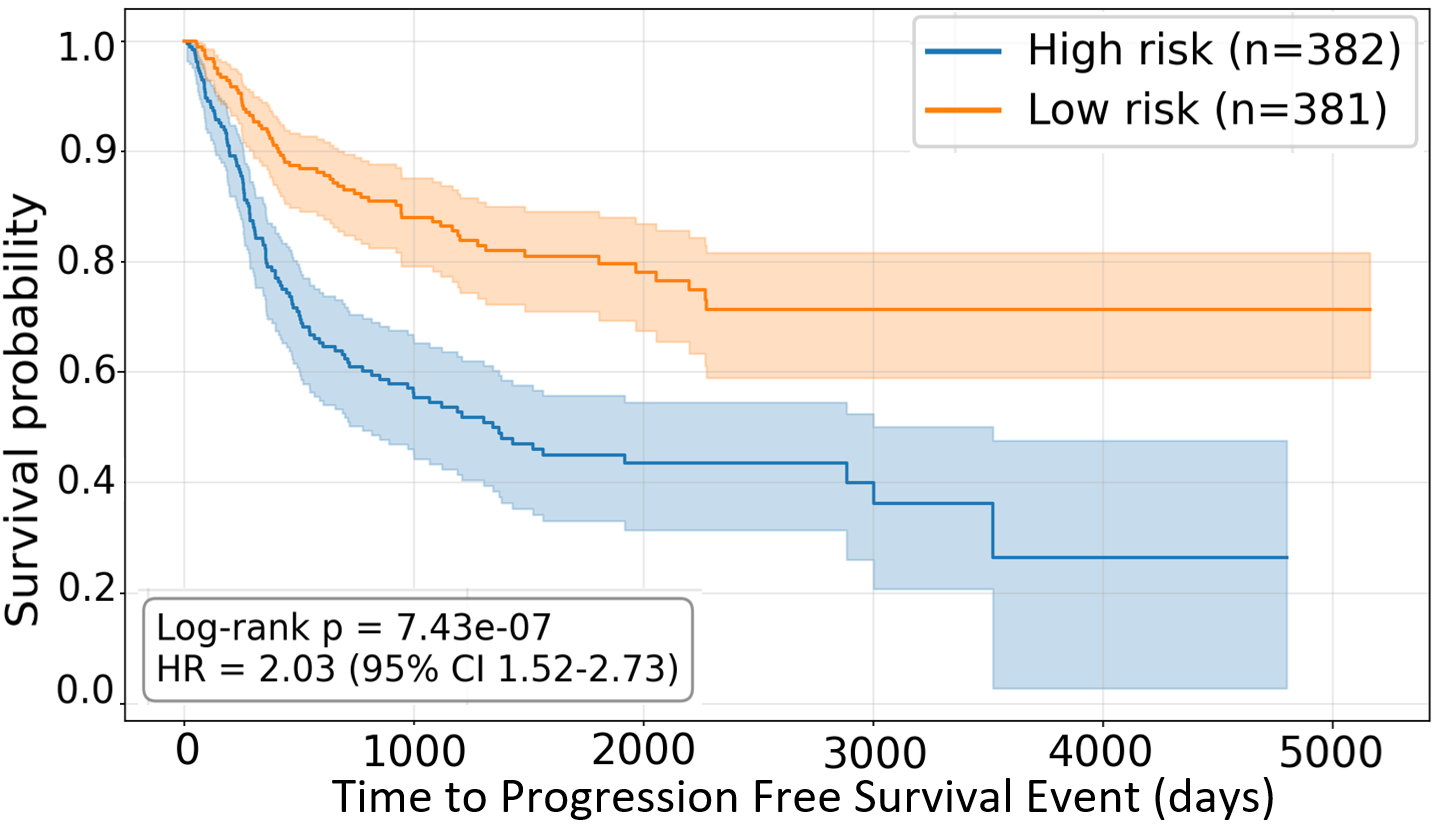} \\

        \methodname{Multi-FRuGaL} &
        \kmplot{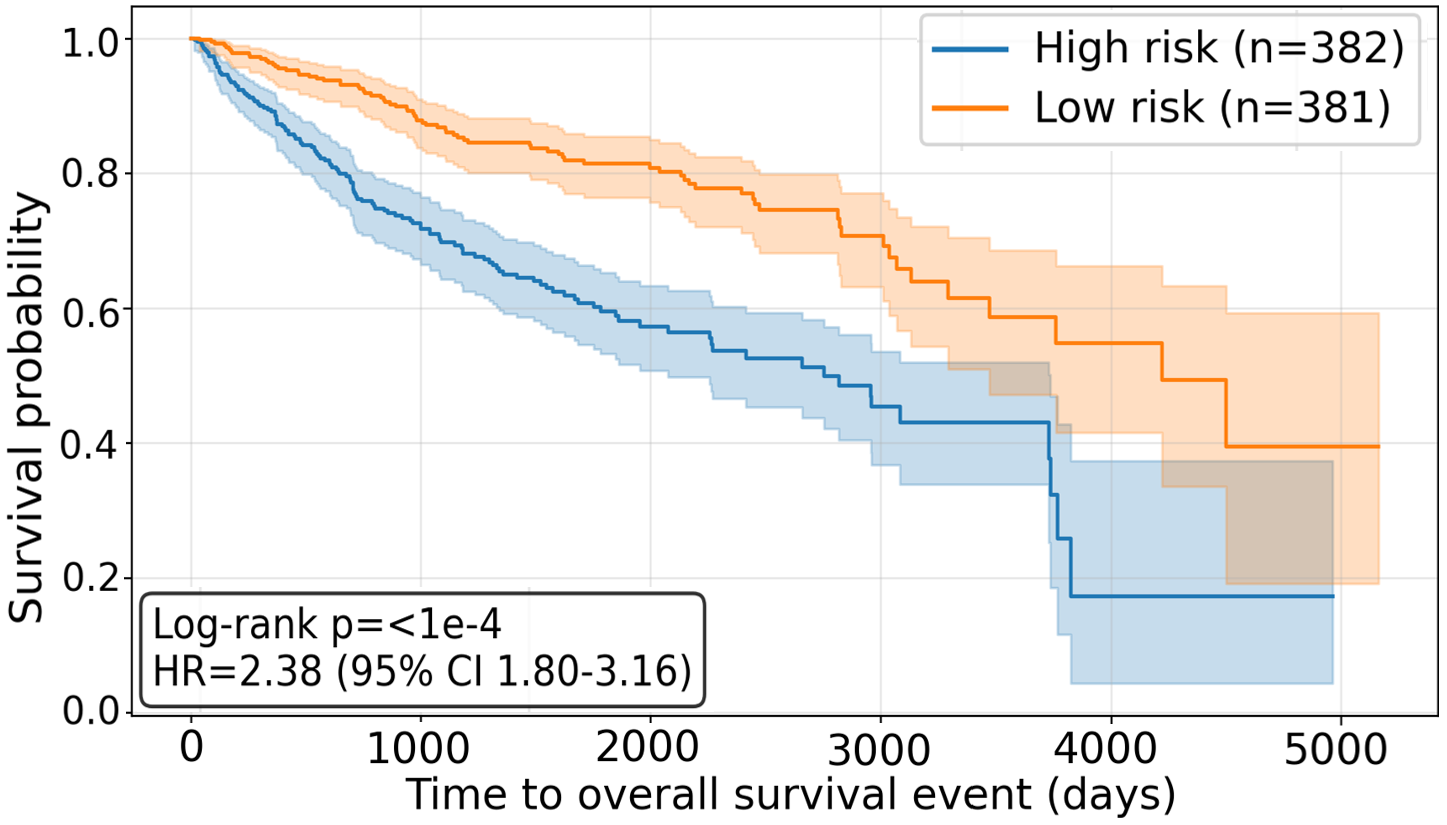} &
        \kmplot{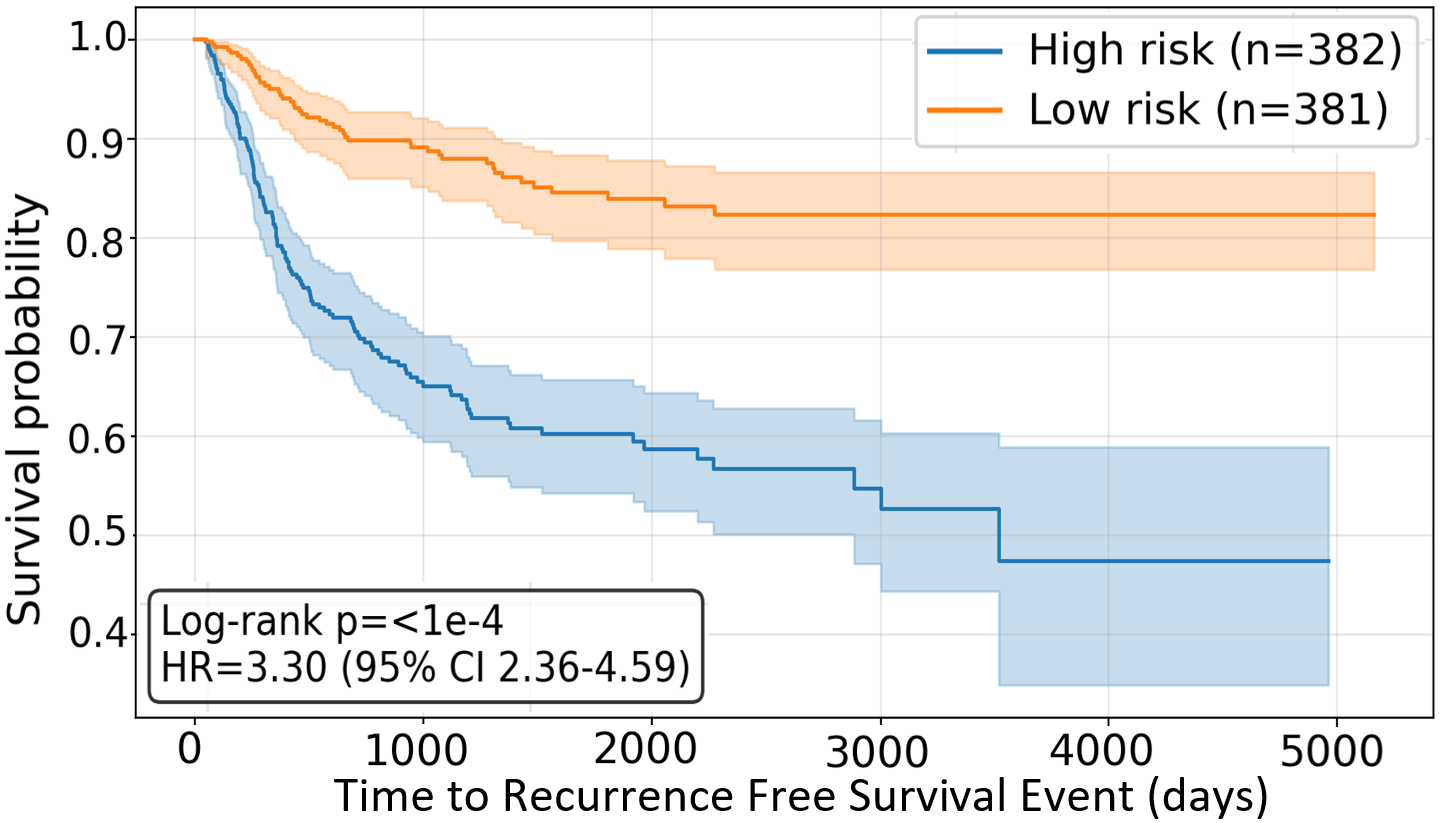} &
        \kmplot{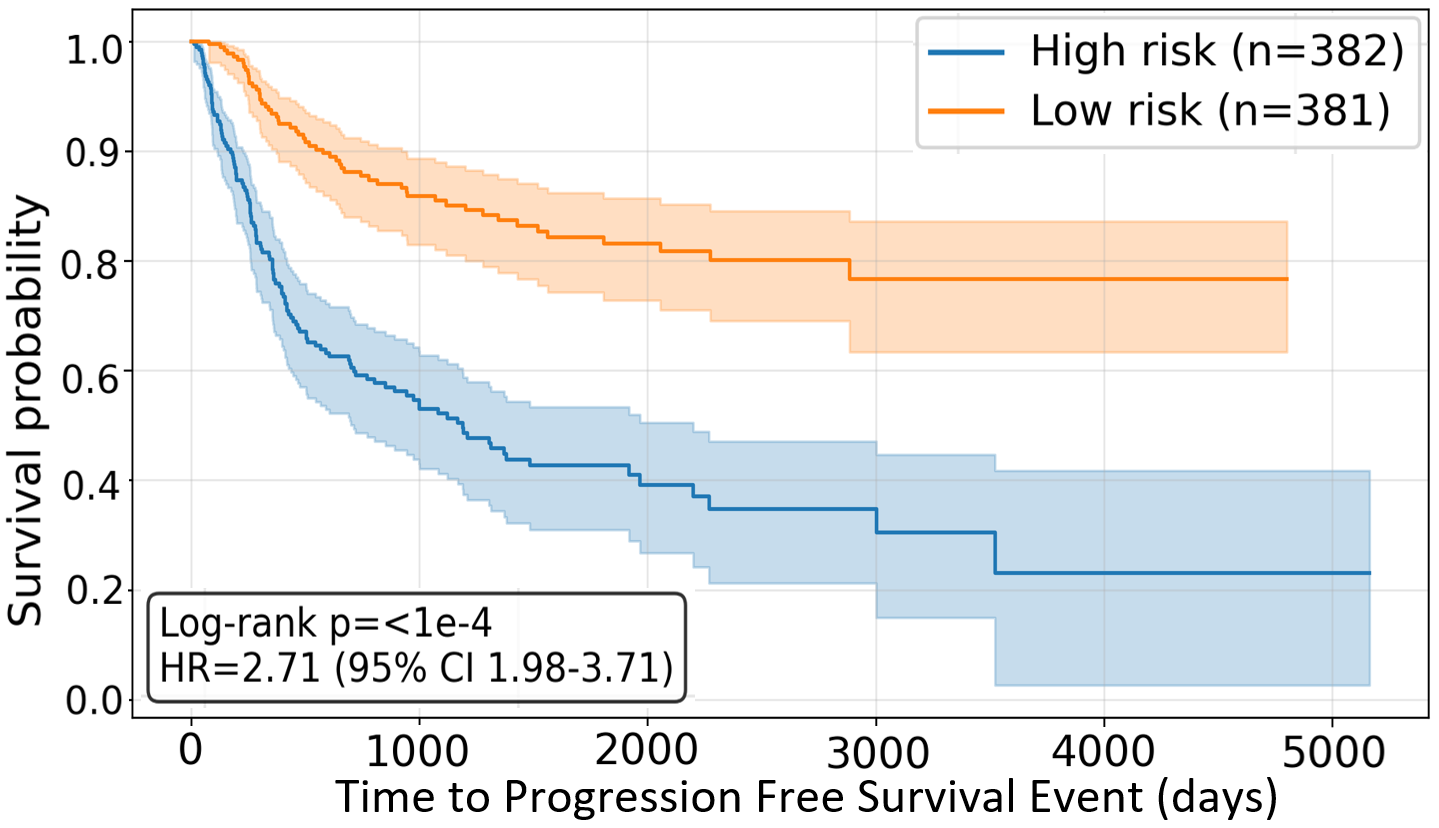} \\

        \bottomrule
    \end{tabular}

    \caption{Kaplan--Meier survival curves on the HANCOCK dataset across competing methods for Overall Survival (OS), Recurrence-Free Survival (RFS), and Progression-Free Survival (PFS). Patients were stratified into high- and low-risk groups using model-predicted risk scores.}
    \label{fig:hancock_km_all_methods}
\end{figure*}

            As shown in Fig.~\ref{fig:hancock_km_all_methods}, the qualitative analysis of the HANCOCK dataset first aligns with the quantitative findings, demonstrating clear survival stratification between the predicted high- and low-risk groups across RFS, OS, and PFS.  Consistent with the reported C-index improvements, the low-risk group maintains systematically higher survival probabilities than the high-risk group across all three endpoints, confirming that the learned representations capture clinically meaningful prognostic structure beyond aggregate numerical metrics. Among these outcomes, RFS and PFS show the clearest and most sustained separation over time, while OS exhibits a more gradual divergence but still preserves meaningful discrimination throughout follow-up. 

            As shown in Fig.~\ref{fig:hancock_km_all_methods}, the qualitative Kaplan--Meier analysis of the HANCOCK dataset is consistent with the quantitative trends reported for HONeYBEE, MulT, LANISTR, and Multi-FRuGaL, demonstrating survival stratification between the predicted high- and low-risk groups across RFS, OS, and PFS. The accompanying log-rank $p$-values and hazard ratios (HRs) with 95\% confidence intervals further quantify the statistical significance and magnitude of this separation. Consistent with the reported C-index improvements, Multi-FRuGaL produces the clearest separation, with the low-risk group maintaining systematically higher survival probabilities than the high-risk group across all three endpoints. In contrast, the comparatively weaker stratification observed for HONeYBEE, MulT, and LANISTR reflects their architectural limitations: feature concatenation treats modality embeddings as a single fixed vector and is therefore sensitive to noisy or high-variance inputs; cross-attention models cross-modal interactions but does not explicitly suppress redundant or unreliable modality contributions; and generative fusion can stabilize missing-modality settings but may propagate uncertainty through imputed latent representations. By contrast, Multi-FRuGaL explicitly decomposes each modality into shared and modality-specific components, applies patient-wise gating, and penalizes redundant information, enabling the fused representation to emphasize complementary prognostic signals while suppressing noisy or duplicated evidence. This decomposition- and gating-driven fusion explains the stronger and more sustained KM separation observed for Multi-FRuGaL, particularly for RFS and PFS, while OS exhibits a more gradual but still meaningful divergence throughout follow-up.
            
\begin{figure*}[t]
    \centering

    \begin{minipage}[t]{0.9\textwidth}
        \centering
        \includegraphics[width=\linewidth]{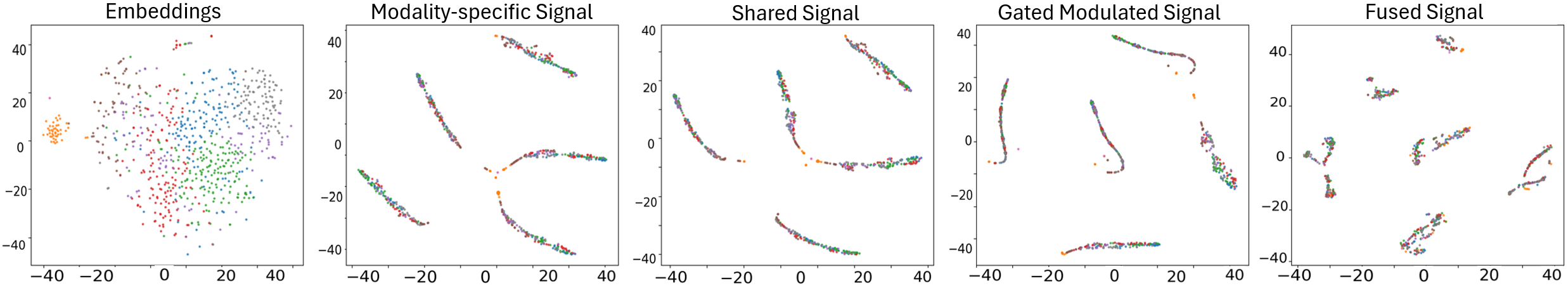}
        
        \small (a) TMA modality
    \end{minipage}

    \vspace{4mm}

    \begin{minipage}[t]{0.9\textwidth}
        \centering
        \includegraphics[width=\linewidth]{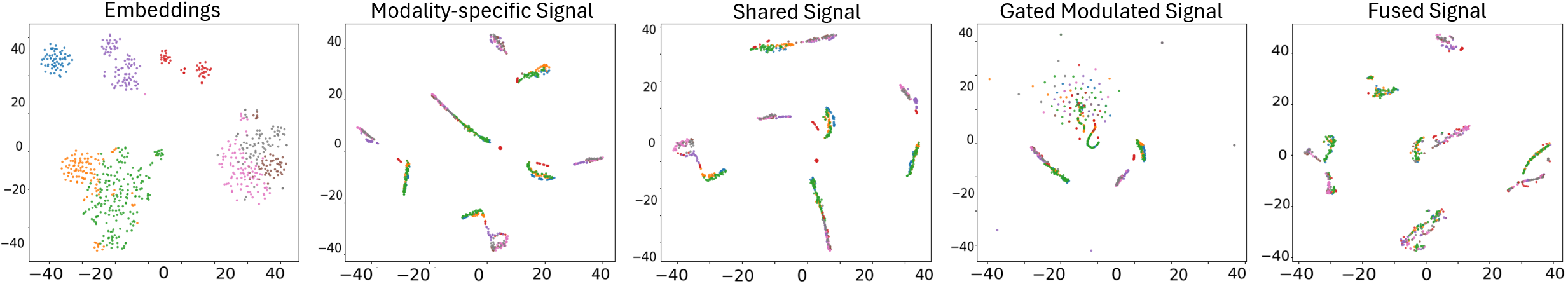}
        
        \small (b) WSI modality
    \end{minipage}

    \vspace{4mm}

    \begin{minipage}[t]{0.9\textwidth}
        \centering
        \includegraphics[width=\linewidth]{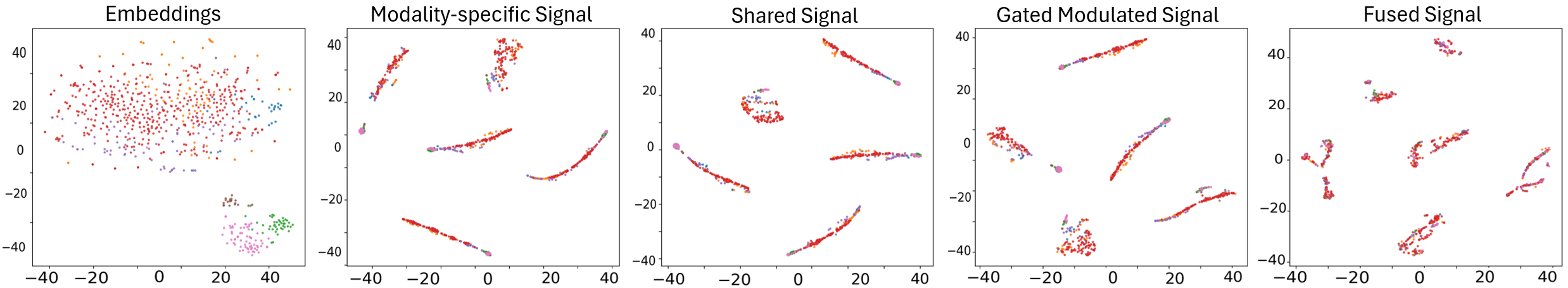}
        
        \small (c) Text modality
    \end{minipage}

    \vspace{4mm}

    \begin{minipage}[t]{0.9\textwidth}
        \centering
        \includegraphics[width=\linewidth]{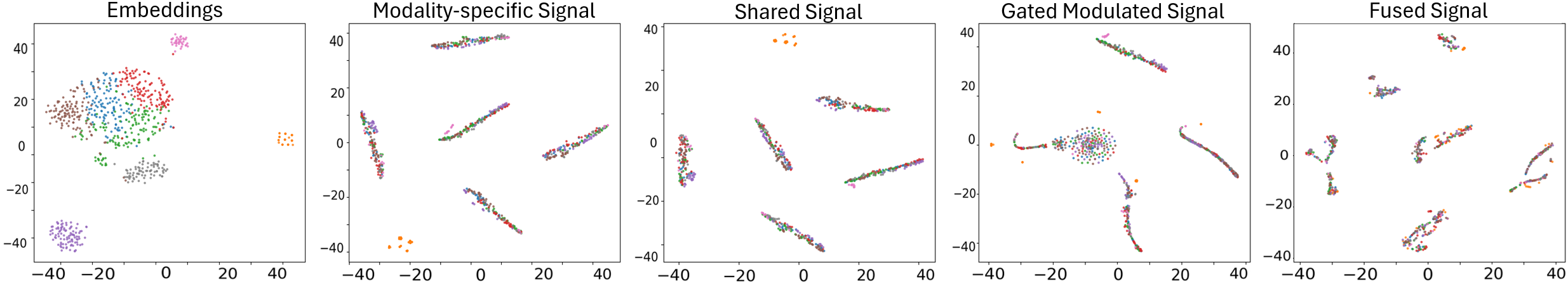}
        
        \small (d) WSI-primary modality
    \end{minipage}

    \vspace{4mm}

    \begin{minipage}[t]{0.9\textwidth}
        \centering
        \includegraphics[width=\linewidth]{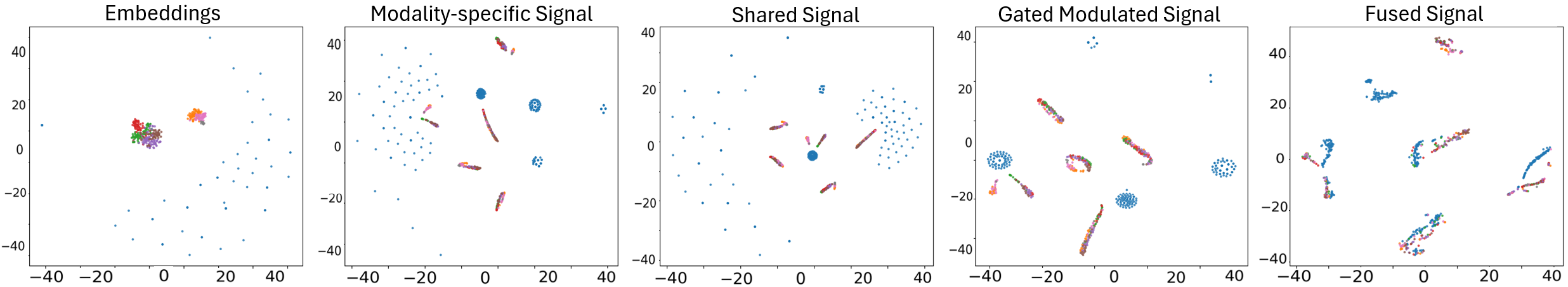}
        
        \small (e) WSI-lymph modality
    \end{minipage}

    \vspace{4mm}

    \begin{minipage}[t]{0.9\textwidth}
        \centering
        \includegraphics[width=\linewidth]{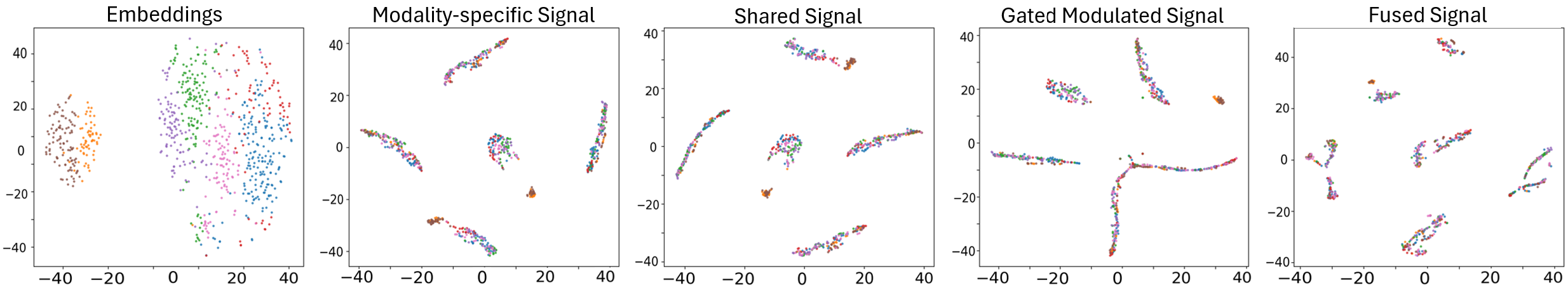}
        
        \small (f) Tabular modality
    \end{minipage}

    \caption{Stage-wise t-SNE visualizations across HANCOCK modalities. Each point represents one patient embedding. Colors denote K-means clusters computed from the raw embedding space of each modality and are kept fixed across stages to visualize how local embedding groups evolve from raw, modality-specific, modality-shared, gated, and fused representations. These projections are intended as qualitative visualization only.}
    \label{fig:hancock_stagewise_tsne}
\end{figure*}

            Building on this survival-level evidence, Fig.~\ref{fig:hancock_stagewise_tsne} provides a stage-wise view of how Multi-FRuGaL organizes multimodal information across tabular, text, TMA, primary-WSI, lymph-node WSI, and combined WSI inputs. In the raw embedding space, the modality representations appear diffuse, weakly organized, and less separable, suggesting that the original features are not sufficiently structured for robust multimodal integration. As the representations pass through the modality-specific and modality-shared decomposition branches, their geometry becomes progressively more compact and organized, indicating that the framework disentangles modality-specific variation from cross-modal disease-relevant structure before fusion. This trend becomes even more evident after gating and in the final fused space, where the representations exhibit the most stable, compact, and globally coherent arrangement across modalities, qualitatively supporting the quantitative gains and suggesting that Multi-FRuGaL learns a more discriminative multimodal latent space under heterogeneous inputs. 

\begin{figure*}[t]
        \centering
        \includegraphics[
            width=0.95\textwidth,
            height=0.30\textheight,
            keepaspectratio
        ]{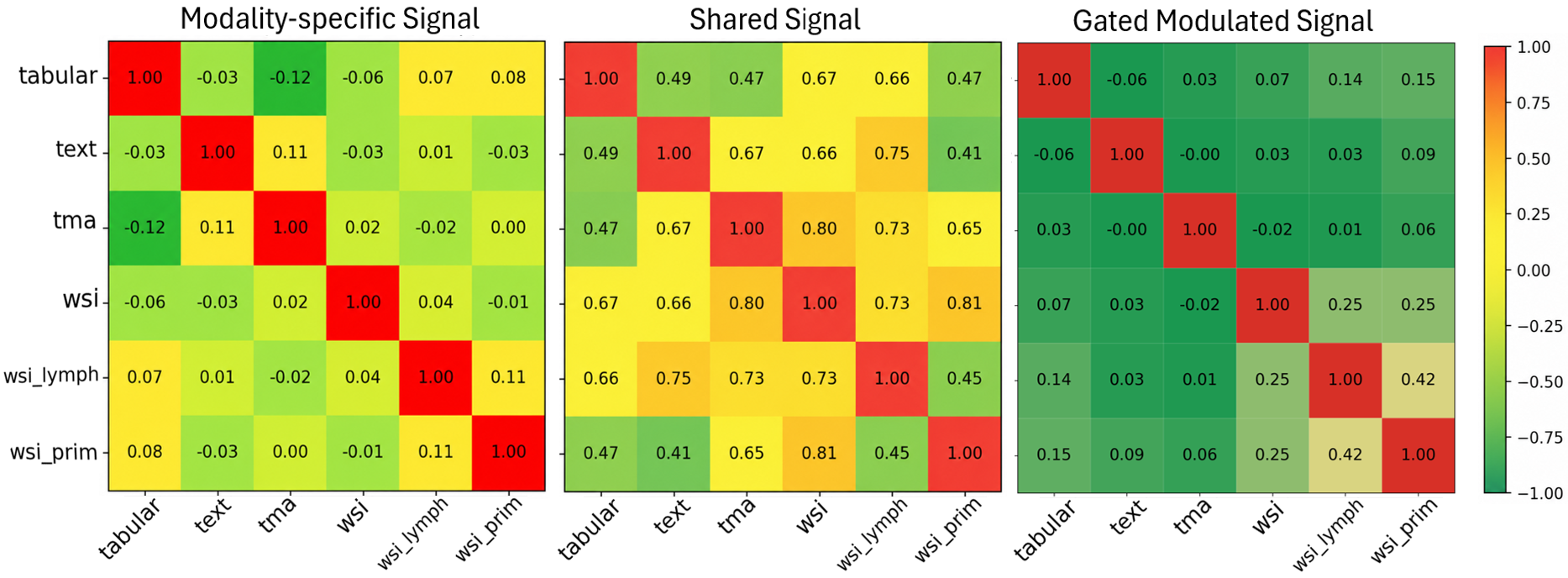}
        \caption{Cross-modal cosine similarity matrices for modality-specific, modality-shared, and gated representations on the HANCOCK dataset. The modality-specific representations show weak or near-zero pairwise similarity, indicating that modality-specific information is largely preserved without redundant duplication. In contrast, the modality-shared representations exhibit moderate to strong cross-modal alignment, consistent with their intended role in capturing disease-relevant information shared across modalities. After gating, the similarities remain structured but moderated, suggesting that the model selectively retains informative cross-modal relationships while suppressing excessive redundancy before final fusion.}
        \label{fig:hancock_cosine_similarity}
    \end{figure*}

            This geometric interpretation is further reinforced by the cosine-similarity analysis in Fig.~\ref{fig:hancock_cosine_similarity}, which shows that the decomposition behaves consistently with its intended design. Red indicates stronger positive cosine similarity, yellow indicates moderate or near-zero similarity, and green indicates weak or negative similarity. Specifically, the modality-shared branch demonstrates stronger cross-modal alignment than the modality-specific branch, indicating that it captures common disease-relevant information preserved across modalities, whereas the modality-specific branch exhibits near-zero or weak pairwise similarities, supporting the claim that modality-specific information is isolated rather than redundantly duplicated. After the gating stage, the similarities remain moderated rather than collapsing into uniformly high agreement, suggesting that the model retains useful shared structure while suppressing excessive redundancy prior to final fusion.
            
\begin{figure*}[t]
    \centering

    \begin{minipage}[t]{0.48\textwidth}
        \centering
        \includegraphics[width=\linewidth]{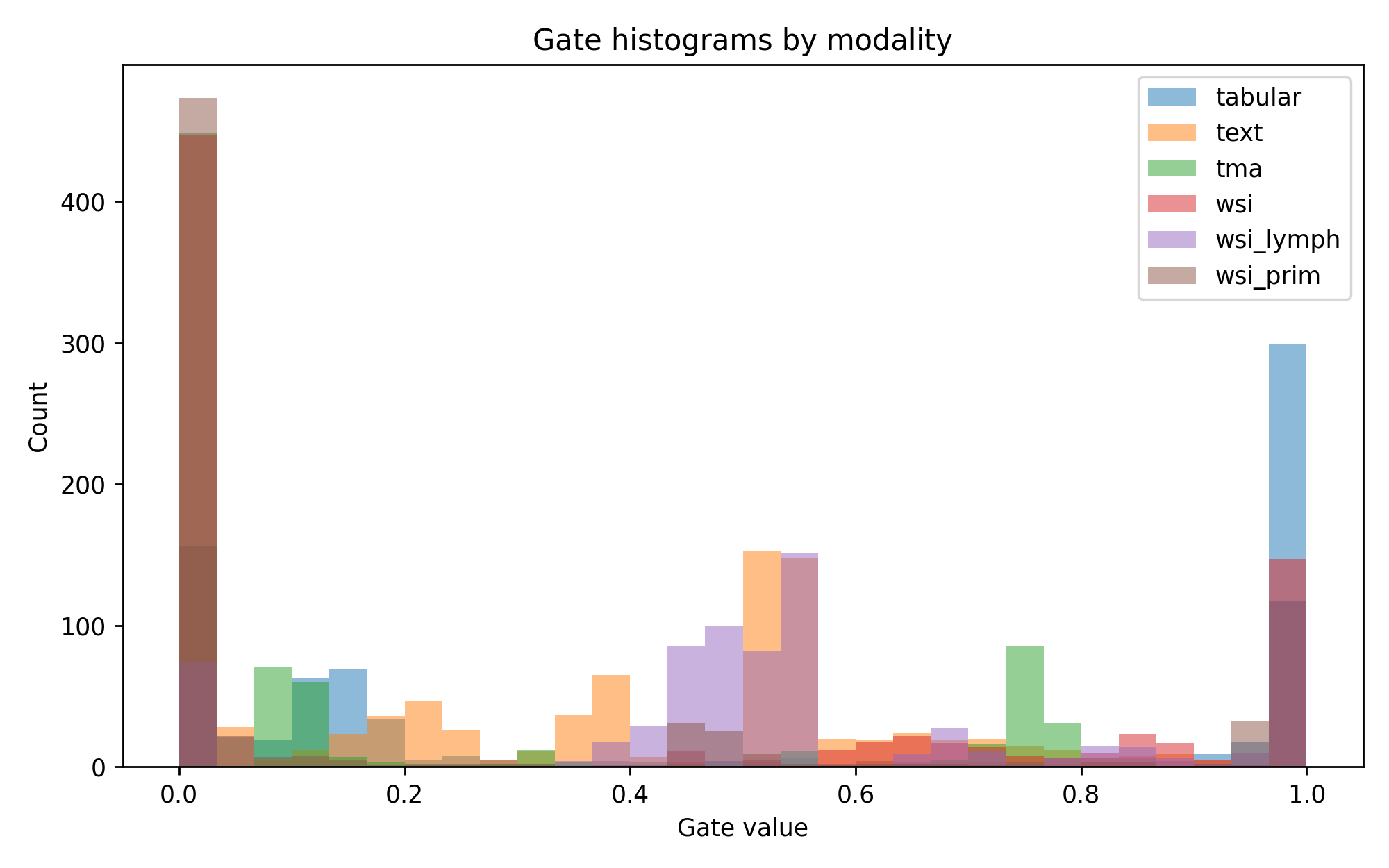}
        
        \small (a) Gate histograms by modality
    \end{minipage}\hfill
    \begin{minipage}[t]{0.48\textwidth}
        \centering
        \includegraphics[width=\linewidth]{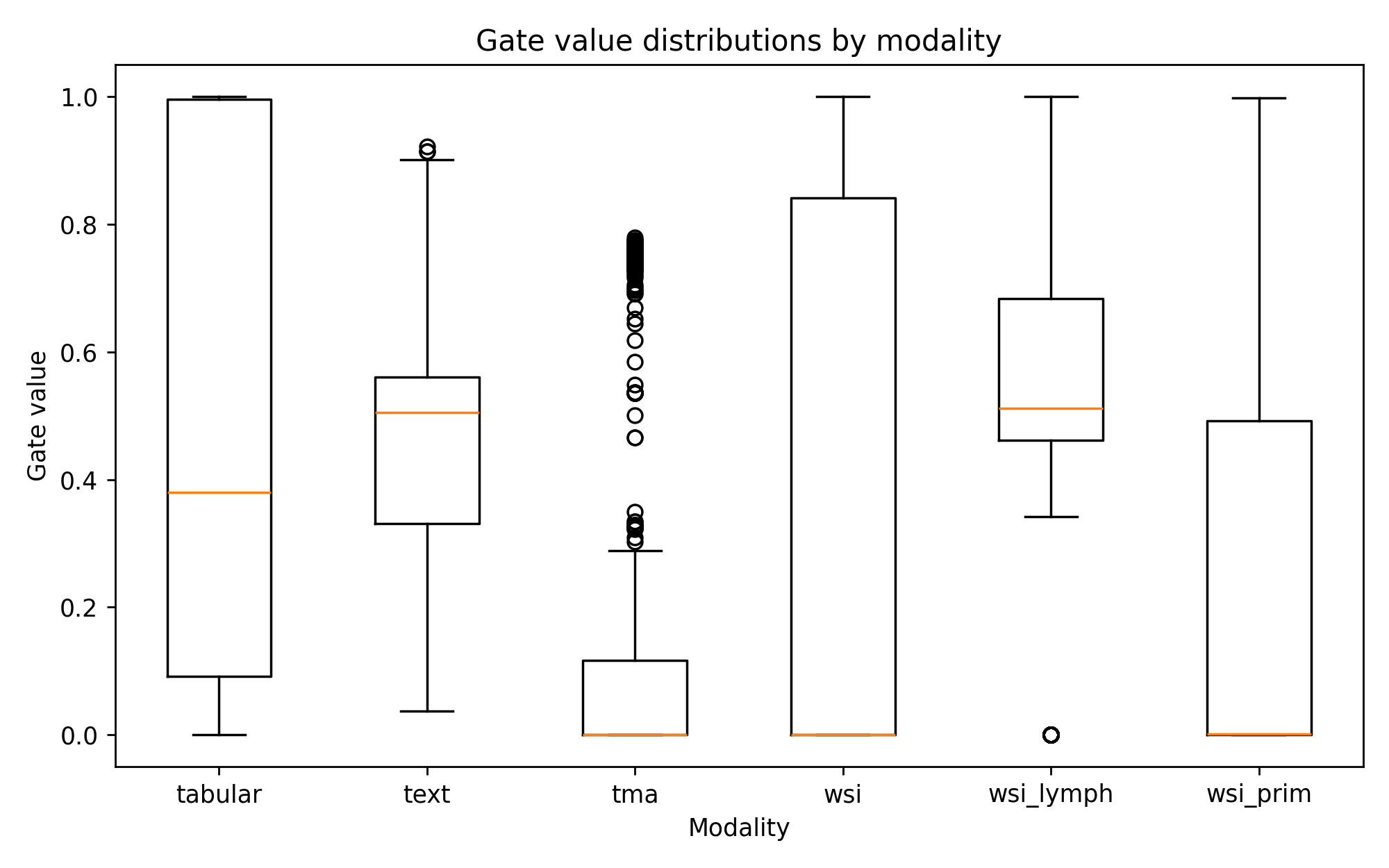}
        
        \small (b) Gate value distributions by modality
    \end{minipage}

    \caption{Distribution of learned gate values across HANCOCK modalities. The histogram view shows the overlap and concentration of gate values, while the boxplot summarizes the central tendency and spread for each modality.}
    \label{fig:hancock gate distributions}
\end{figure*}

            Finally, Fig.~\ref{fig:hancock gate distributions} offers patient-level evidence for the adaptive behavior of GateNet by showing the distribution of learned modality gates across the HANCOCK cohort. Rather than assigning a single fixed importance to each modality, the broad spread of gate values indicates that modality contributions are dynamically adjusted on a sample-by-sample basis, which is especially important in the presence of heterogeneous and incomplete multimodal inputs. In particular, tabular, text, and lymph-node WSI features more frequently receive higher gate activations, suggesting that they contribute more consistently to downstream prediction, whereas TMA and primary-WSI display wider variability and more frequent suppression, implying that their utility is more context-dependent and patient-specific. Taken together, these qualitative results form a coherent progression from outcome-level stratification, to latent-space organization, to decomposition fidelity, and finally to adaptive modality weighting, collectively reinforcing the quantitative results and supporting the conclusion that Multi-FRuGaL improves multimodal prognosis by learning structured, non-redundant, and dynamically weighted representations.

        \subsubsection{HECKTOR Dataset}
\begin{figure*}[t]
    \centering

    \newcommand{\hkmw}{0.45\textwidth}
    \newcommand{\hkmh}{0.24\textheight}

    \begin{tabular}{c c}
        \toprule
        \textbf{HONeYBEE} & \textbf{MulT} \\
        \midrule
        \includegraphics[width=\hkmw,height=\hkmh,keepaspectratio]{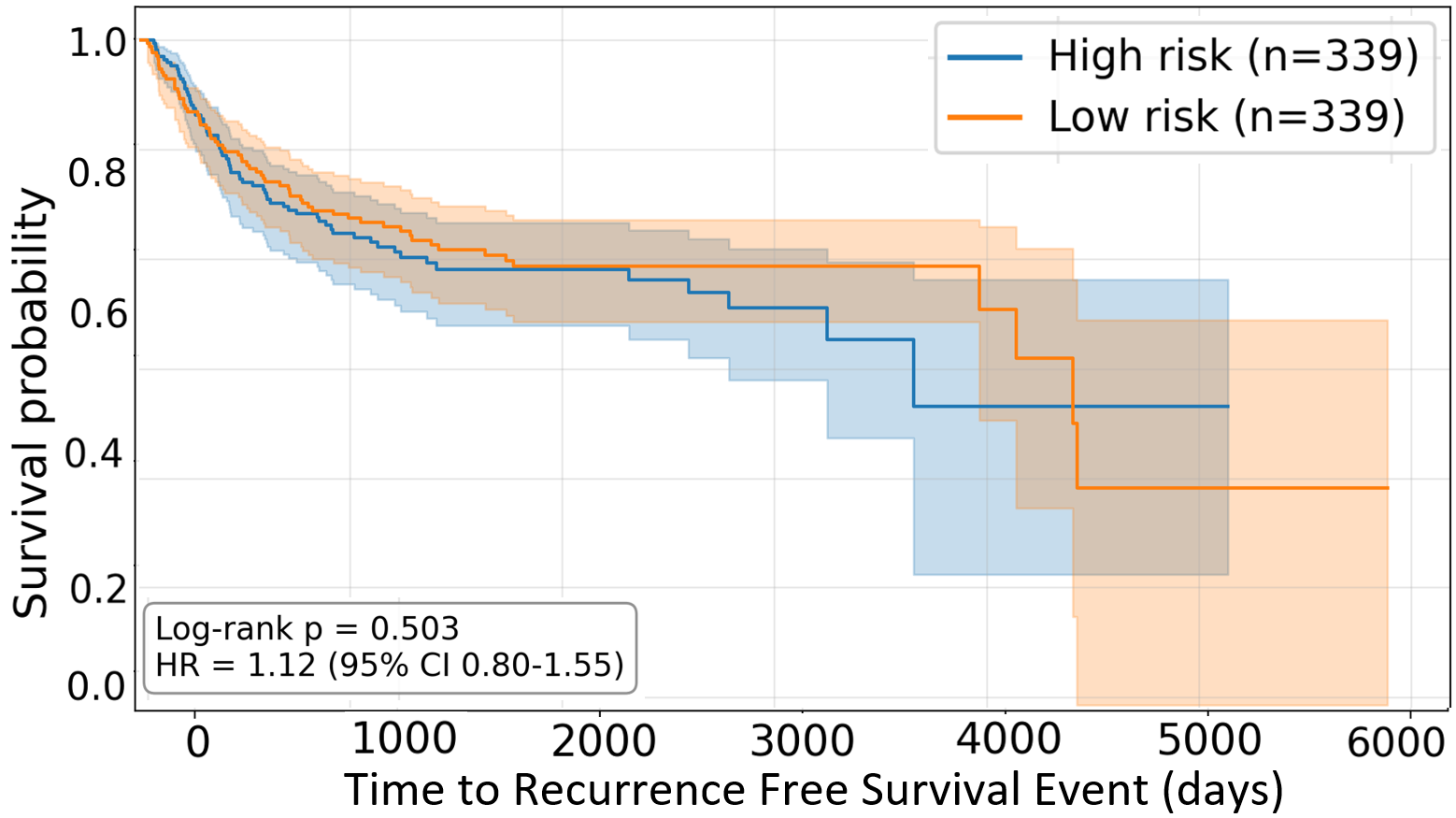} &
        \includegraphics[width=\hkmw,height=\hkmh,keepaspectratio]{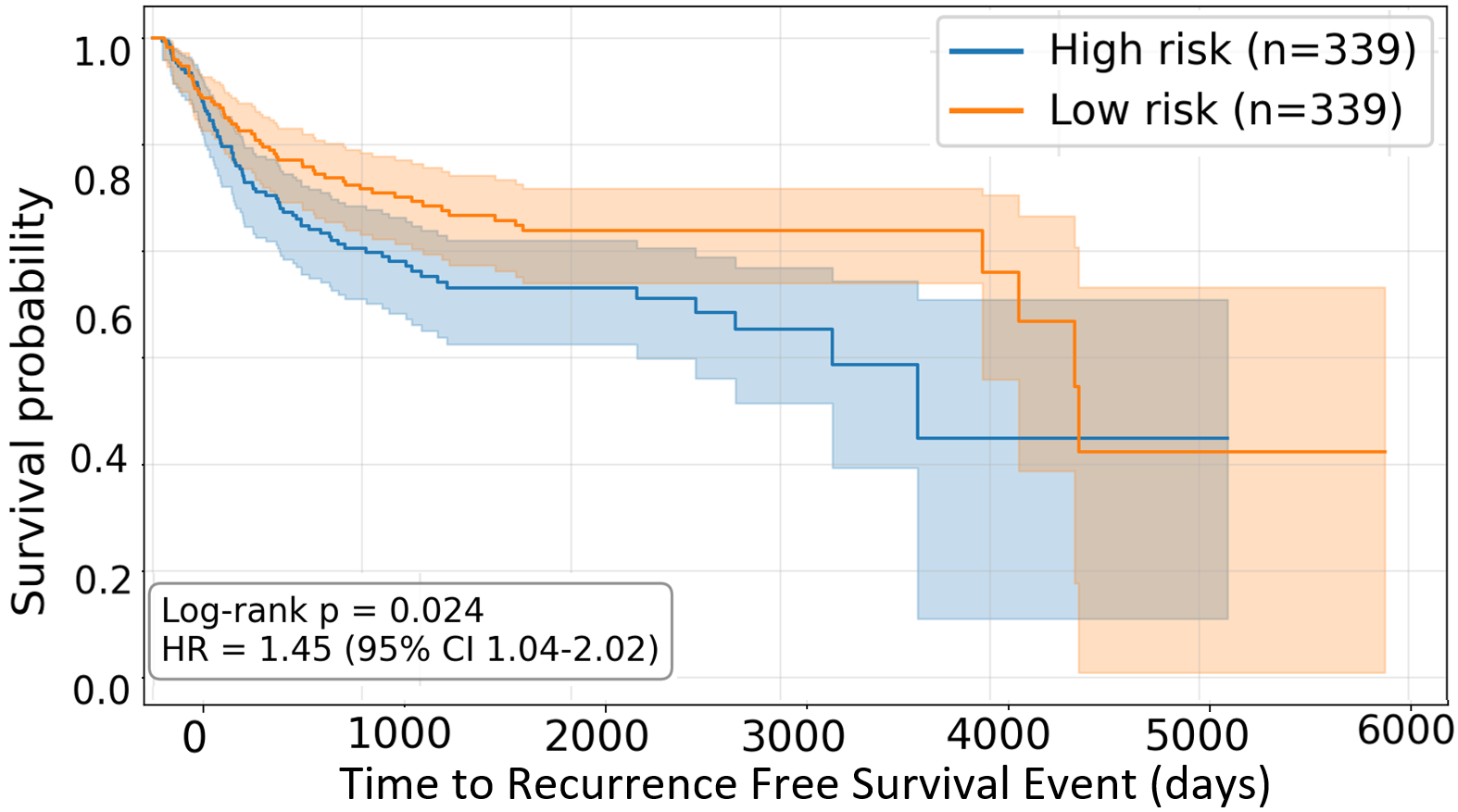} \\

        \midrule
        \textbf{LANISTR} & \textbf{Multi-FRuGaL} \\
        \midrule
        \includegraphics[width=\hkmw,height=\hkmh,keepaspectratio]{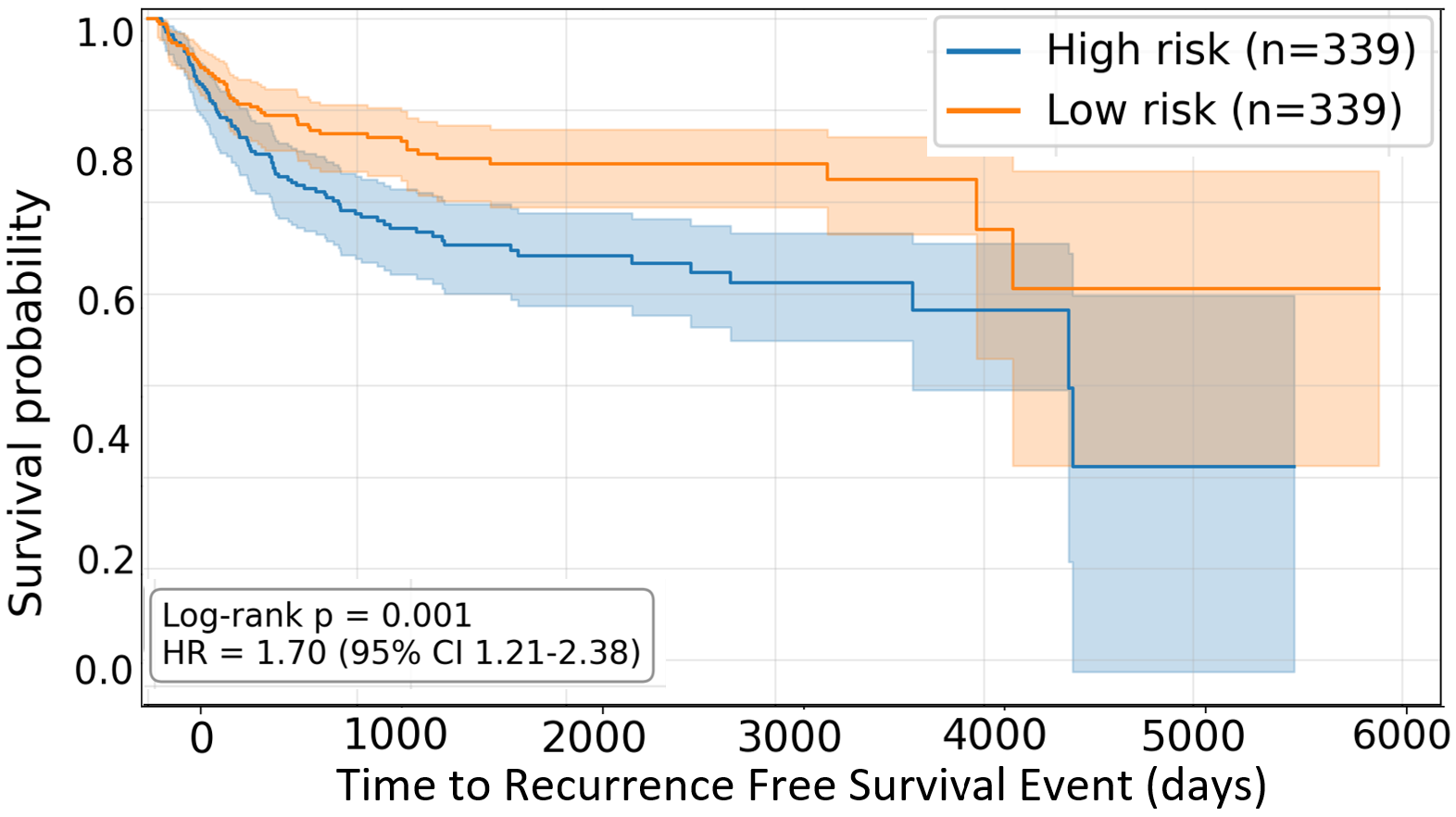} &
        \includegraphics[width=\hkmw,height=\hkmh,keepaspectratio]{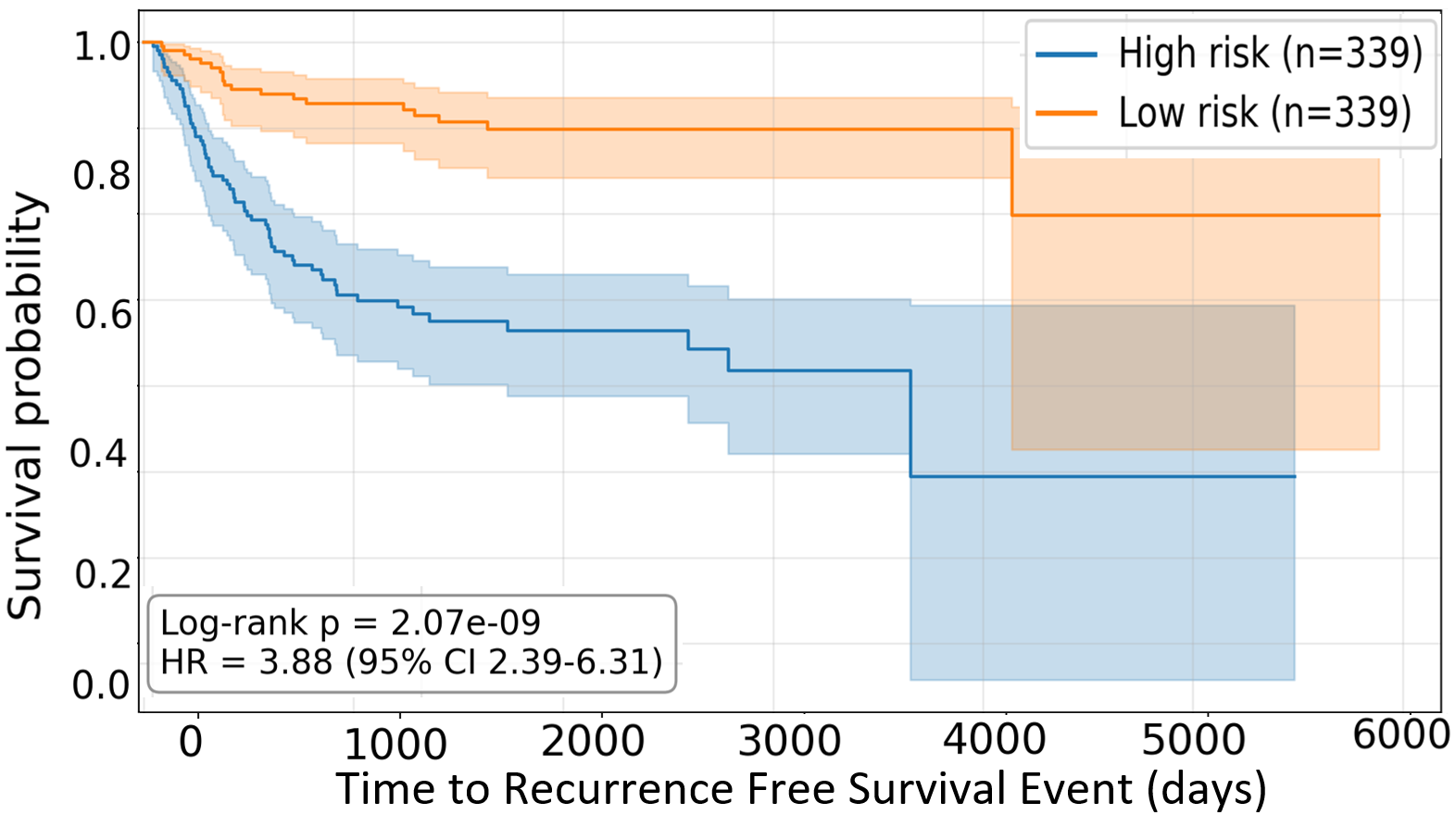} \\

        \bottomrule
    \end{tabular}

    \caption{Kaplan--Meier survival curves on the HECKTOR dataset for recurrence-free survival (RFS) across competing methods. Patients were stratified into high- and low-risk groups using model-predicted risk scores.}
    \label{fig:hecktor_km_all_methods}
\end{figure*}

\begin{figure*}[t]
    \centering

    \begin{minipage}[t]{0.9\textwidth}
        \centering
        \includegraphics[width=\linewidth]{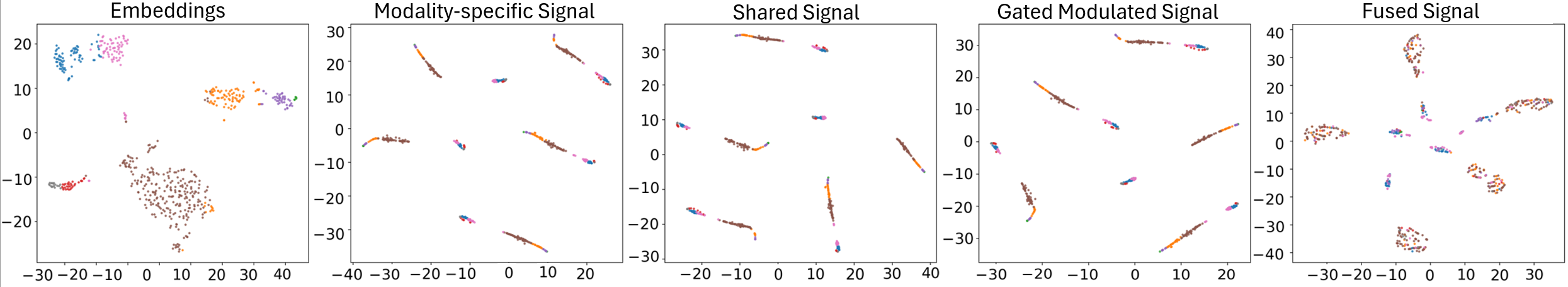}
        
        \small (a) CT modality
    \end{minipage}

    \vspace{4mm}

    \begin{minipage}[t]{0.9\textwidth}
        \centering
        \includegraphics[width=\linewidth]{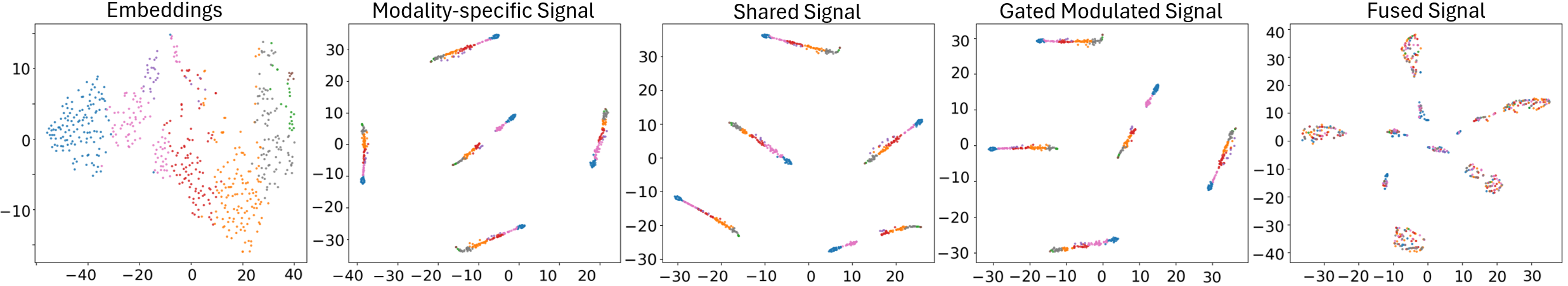}
        
        \small (b) PET modality
    \end{minipage}

    \vspace{4mm}

    \begin{minipage}[t]{0.9\textwidth}
        \centering
        \includegraphics[width=\linewidth]{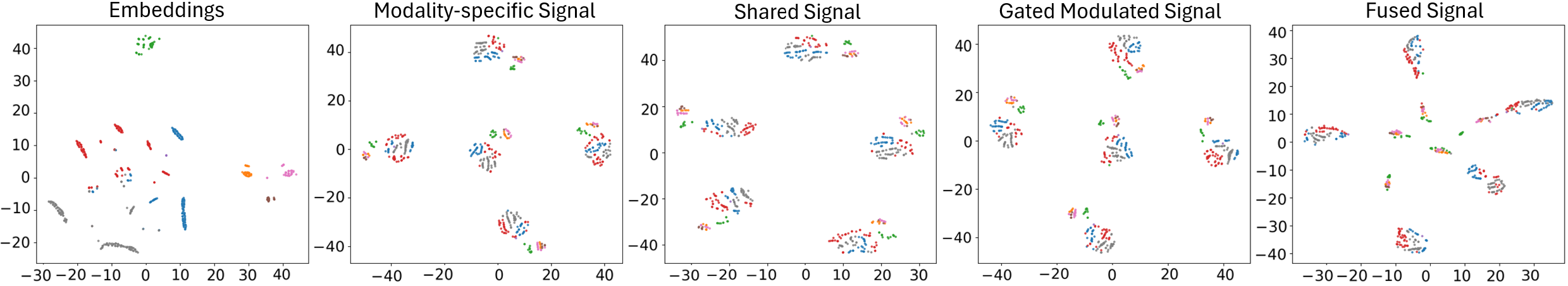}
        
        \small (c) Tabular modality
    \end{minipage}

    \caption{Stage-wise t-SNE visualizations across HECKTOR modalities. For each modality, the five panels correspond to the raw, modality-specific, modality-shared, gated, and fused representations, respectively.}
    \label{fig:hecktor_stagewise_tsne}
\end{figure*}

\begin{figure*}[t]
    \centering

    \includegraphics[
        width=0.95\textwidth,
        height=0.30\textheight,
        keepaspectratio
    ]{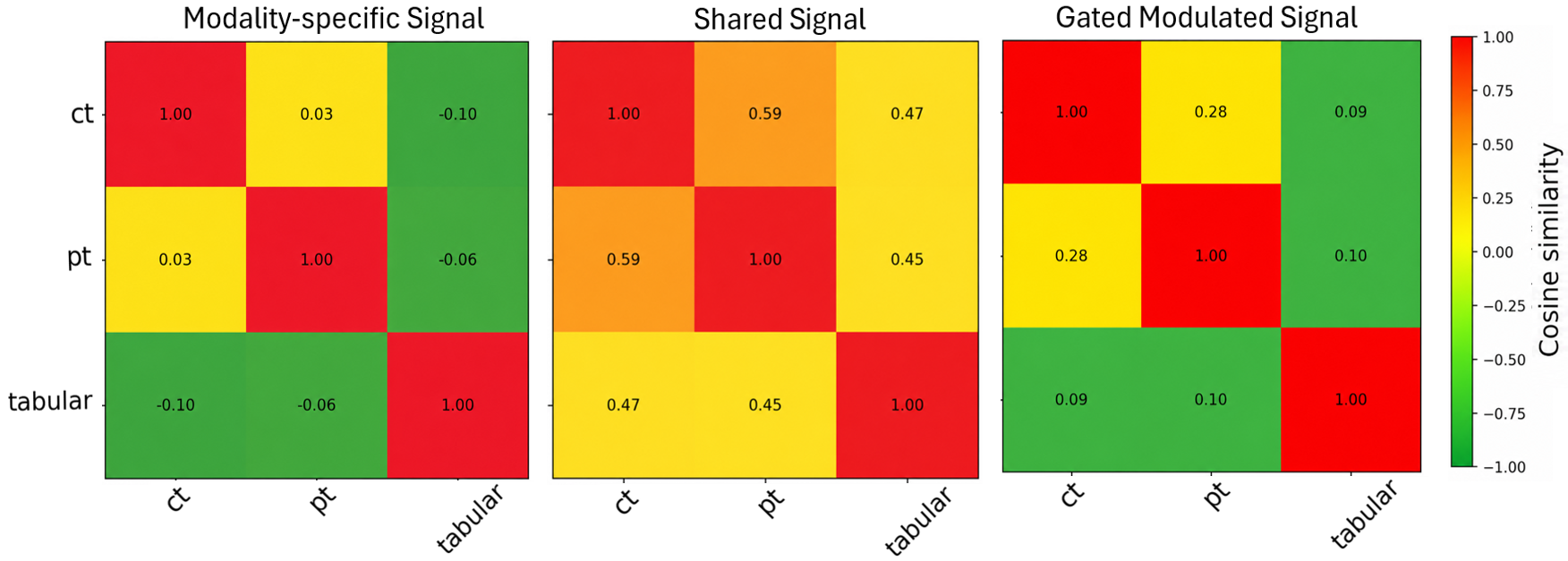}

\caption{Cross-modal cosine similarity matrices for modality-specific, modality-shared, and gated representations on the HECKTOR dataset. The modality-specific representations show near-zero or weak pairwise similarity, indicating that modality-specific information is preserved with limited cross-modal overlap. In contrast, the modality-shared representations exhibit moderate positive cross-modal alignment, consistent with the intended role of the shared branch in capturing common disease-relevant information across CT, PET, and tabular modalities. After gated modulation, cross-modal similarities are reduced relative to the shared branch but remain positive, suggesting that the gating mechanism retains useful shared structure while suppressing excessive redundancy before fusion. }
\label{fig:hecktor_cosine_similarity}
\end{figure*}

\begin{figure*}[t]
    \centering

    \begin{minipage}[t]{0.48\textwidth}
        \centering
        \includegraphics[width=\linewidth]{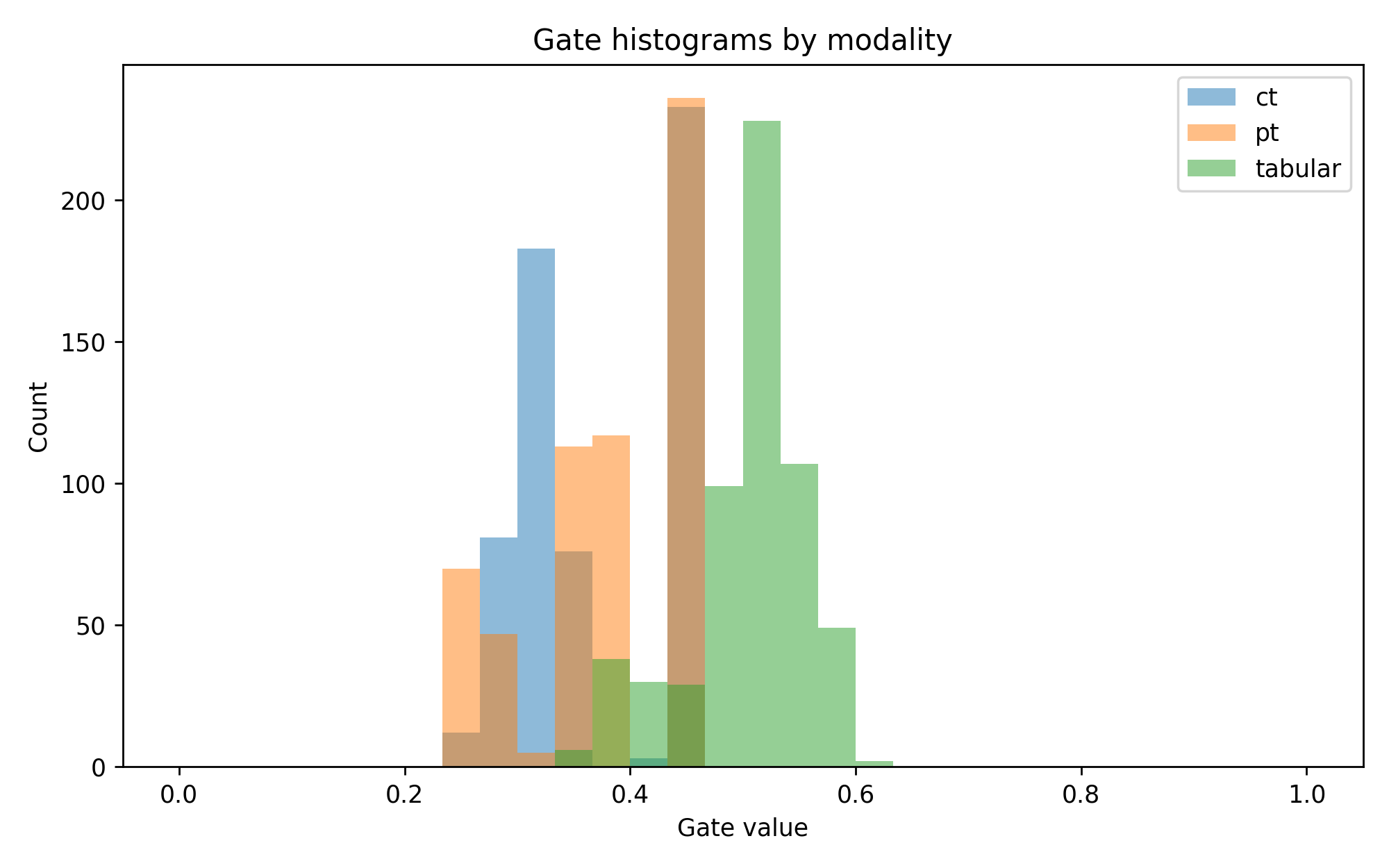}
        
        \small (a) Gate histograms by modality
    \end{minipage}\hfill
    \begin{minipage}[t]{0.48\textwidth}
        \centering
        \includegraphics[width=\linewidth]{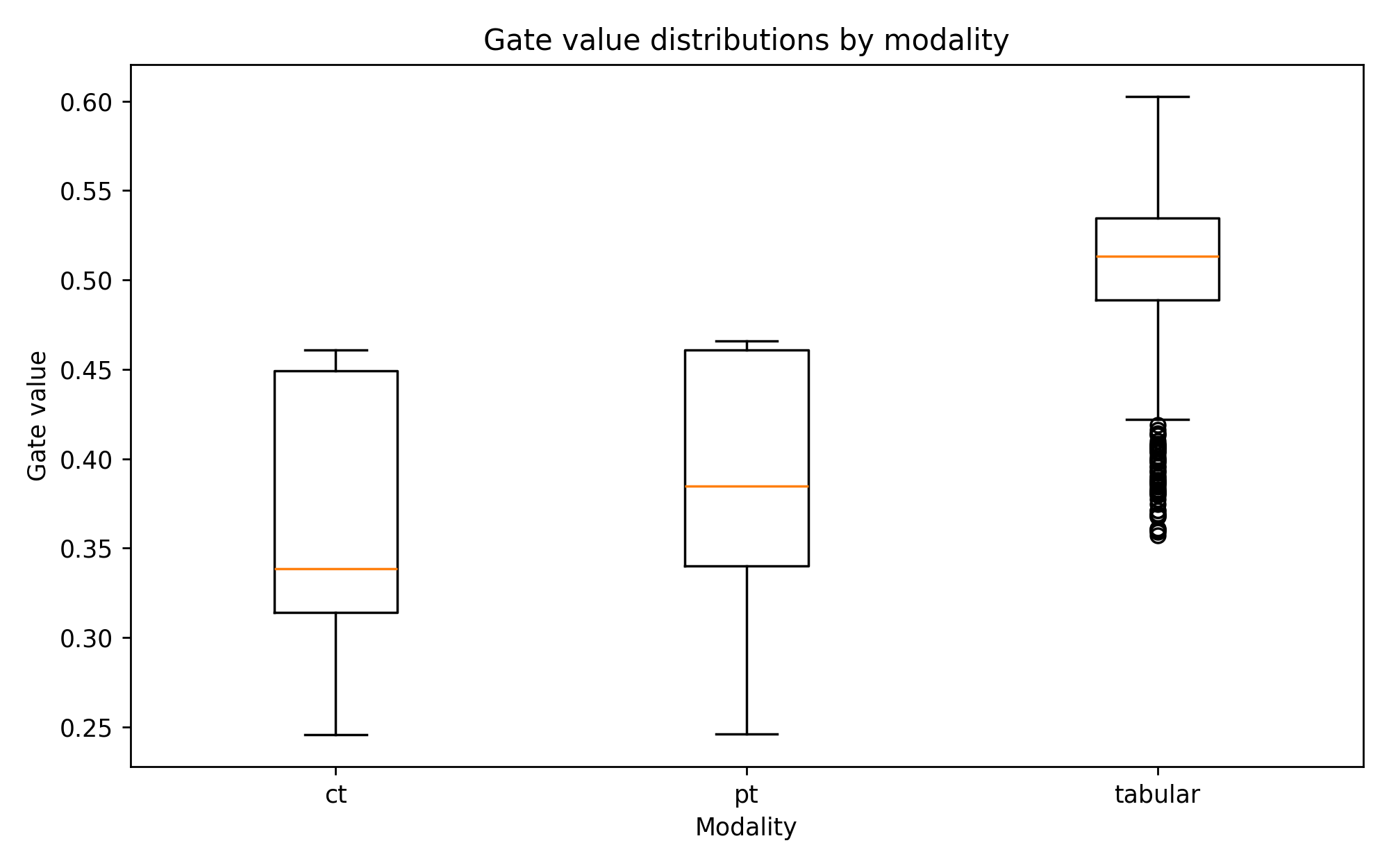}
        
        \small (b) Gate value distributions by modality
    \end{minipage}

    \caption{Distribution of learned gate values across HECKTOR modalities. The histogram view shows the overlap and concentration of gate values, while the boxplot summarizes the central tendency and spread for each modality.}
    \label{fig:hecktor_gate_distributions}
\end{figure*}

            As shown across Figs.~\ref{fig:hecktor_km_all_methods}, \ref{fig:hecktor_stagewise_tsne}, \ref{fig:hecktor_cosine_similarity}, and \ref{fig:hecktor_gate_distributions}, the qualitative results on the HECKTOR dataset form a coherent picture that closely aligns with the quantitative improvements. The Kaplan--Meier analysis in Fig.~\ref{fig:hecktor_km_all_methods} is consistent with the C-index trends observed across HONeYBEE, MulT, LANISTR, and Multi-FRuGaL, with Multi-FRuGaL showing the strongest RFS stratification between predicted high- and low-risk groups. The accompanying log-rank $p$-values and hazard ratios (HRs) with 95\% confidence intervals further quantify the statistical significance and magnitude of this separation. Compared with the weaker separation observed for feature concatenation, cross-attention, and generative fusion baselines, the sustained divergence of the Multi-FRuGaL curves suggests that its decomposition-aware gated fusion better captures clinically meaningful risk structure rather than only improving aggregate ranking metrics. This outcome-level separation is supported by the representation analysis in Fig.~\ref{fig:hecktor_stagewise_tsne}, where the t-SNE projections show a clear evolution of tabular, CT, and PET embeddings from the raw space to the modality-specific, modality-shared, gated, and final fused stages. In the raw space, the modality-specific embeddings are comparatively diffuse and unevenly organized, whereas the modality-specific and modality-shared branches yield cleaner and more compact geometric patterns, suggesting that the decomposition stage successfully separates modality-specific variation from cross-modal common structure.

            After gating, the embeddings remain structured without collapsing into a single overly uniform manifold, indicating that the model preserves complementary information while reducing unnecessary redundancy. Most notably, the fused space exhibits the most coherent and consistently separated global arrangement across all three modalities, qualitatively reinforcing the quantitative gains and suggesting that Multi-FRuGaL learns a more organized and discriminative multimodal latent space for HECKTOR. This interpretation is further corroborated by the cosine-similarity matrices in Fig.~\ref{fig:hecktor_cosine_similarity}, which provide direct evidence that the decomposition behaves as intended. 
            
            In the cosine-similarity matrices, red indicates stronger positive cosine similarity, yellow indicates moderate or near-zero similarity, and green indicates weak or negative similarity. In the modality-shared branch, cross-modal similarities are consistently moderate and positive, with CT--PET showing the highest similarity (0.59), followed by CT--tabular (0.47) and PET--tabular (0.45), indicating that this branch captures disease-relevant information that is common across modalities. In contrast, the modality-specific branch shows near-zero to slightly negative pairwise similarities, supporting effective isolation of modality-specific, non-overlapping information rather than redundant duplication. After gating, the cross-modal similarities are further reduced but remain positive, suggesting that the model suppresses excess redundancy while retaining a controlled amount of complementary cross-modal signal before fusion. 
            
            Finally, the gate histograms and boxplots in Fig.~\ref{fig:hecktor_gate_distributions} show that modality contributions are not fixed globally but are adaptively adjusted on a patient-specific basis. Tabular features receive the highest gate values overall, with a higher median and a distribution concentrated around larger weights, indicating that clinical variables contribute more consistently across the cohort, whereas CT and PET show lower median values and broader overlap, implying that their utility is more variable and context-dependent. Taken together, these qualitative results trace a consistent progression from clinically meaningful risk stratification, to improved latent-space organization, to well-behaved modality-shared and modality-specific decomposition, and finally to adaptive modality weighting, collectively strengthening the quantitative evidence that Multi-FRuGaL improves multimodal learning through structured, non-redundant, and dynamically weighted representations.

\section{Conclusion}
    
    In this study, we addressed the challenge of fusing heterogeneous multimodal data for head and neck cancer prognosis and diagnosis, when patient data contain substantial and irregular missingness. We introduced the \emph{\textbf{Multi}modal \textbf{F}lexible Red\textbf{u}ndancy-aware decomposed \textbf{Ga}ted \textbf{L}earning} (Multi-FRuGaL) as a decomposition-aware gated intermediate-fusion framework that first separates modality-shared and modality-specific signals, and then uses input-conditioned stochastic gates for patient-wise selection of informative modalities and suppression of redundant, colinear, noisy, or conflicting input streams. To further regulate modality usage, Multi-FRuGaL incorporates an information-aware fusion objective that combines a decomposition loss, a sparsity-inducing budget penalty and a redundancy penalty, enabling complementary (rather than overlapping information) to drive prediction.

    Across both the HANCOCK and HECKTOR datasets, Multi-FRuGaL demonstrated large and consistent improvements over concatenation-, cross-attention-, and generative-based fusion baselines. On HANCOCK, Multi-FRuGaL increased AUC from 0.601 to 0.8496 for 5-year OS and from 0.672 to 0.801 for 2-year recurrence prediction, representing absolute gains of 23.2\% and 12.9\%, respectively, over the strongest baseline. On HECKTOR, Multi-FRuGaL achieved an HPV status classification AUC of 0.975, outperforming feature concatenation, cross-attention, and CNN-MLP baselines while substantially improving specificity and balanced accuracy. Ablation analyses confirmed that the signal decomposition, modality-level gating mechanism and the information-budget regularizer are essential contributors to these gains. Under simulated random modality dropout, Multi-FRuGaL degraded smoothly, maintaining competitive performance even when up to 60\% of available modalities were removed. Although this provides a practical patient-level robustness stress test, future work should evaluate structured missingness patterns arising from site-specific acquisition protocols, treatment pathways, or modality-specific clinical workflows. Beyond predictive accuracy, the learned modality gates provide interpretable, patient-specific information allocation profiles, offering insights into how different data sources contribute to each clinical prediction.

    Overall, Multi-FRuGaL provides a decomposition-aware and robust framework for multimodal fusion under real-world data missingness, with potential for extension to higher-dimensional molecular data and further evaluation toward clinical decision-support applications.

\newcommand{\customheading}[1]{%
  \vspace{1em}\noindent\textbf{ #1}\par\nobreak\vspace{0.5em}
}

\customheading{CRediT authorship contribution statement}
\textbf{S.Kachole:} Conceptualization, Methodology, Software, Validation, Formal analysis, Investigation, Writing - original draft, Writing - Review \& Editing, Visualization, Project administration.
\textbf{S.Thakur:} Software, Resources, Data Curation, Writing - Review \& Editing.
\textbf{S.Innani:} Software, Resources, Writing - Review \& Editing.
\textbf{S.Adap:} Software, Resources, Data Curation, Writing - Review \& Editing.
\textbf{S.You:} Conceptualization, Formal analysis, Investigation, Resources, Writing - Review \& Editing.
\textbf{C.Pitarch-Abaigar:} Conceptualization, Software, Resources, Visualization, Writing - Review \& Editing.
\textbf{S.Bakas:} Conceptualization, Resources, Visualization, Supervision, Project administration, Funding acquisition, Writing - Review \& Editing.

\section*{{Funding and Acknowledgments}}{
Research reported in this publication was partially supported by the Informatics Technology for Cancer Research (ITCR) funding program of the National Cancer Institute (NCI) of the National Institutes of Health (NIH), under award number U24CA279629. Computational resources used in this research were partially supported by Lilly Endowment, Inc., through its support for the Indiana University Pervasive Technology Institute. The content of this publication is solely the responsibility of the authors and does not represent the official views of the NIH, or any other funding body.\\
\\
The authors would like to thank the MICCAI HANCOTHON and HECKTOR 2025 challenge organizers for the valuable data resources and the benchmarking environment they made publicly available.}

\section*{Data availability}
The HANCOCK and HECKTOR datasets used in this study are available through their respective MICCAI challenge data access mechanisms and are subject to the corresponding data-use agreements. Patient-level data and derived features cannot be redistributed by the authors due to dataset licensing, privacy, and data-use restrictions.

\section*{Code availability}
The source code for Multi-FRuGaL is available at: \href{https://github.com/IUCompPath/Multi_Frugal/}{GitHub}

\section*{Ethics and patient privacy}
This study used de-identified data made available through the HANCOCK and HECKTOR challenge data access mechanisms. No patient-identifying information is included in this manuscript, figures, tables, or supplementary materials. Data were used in accordance with the applicable challenge data-use terms and institutional requirements.

\section*{{Declaration of competing interest}}
The authors declare that they have no known competing financial interests or personal relationships that could have appeared to influence the work reported in this paper.








\bibliographystyle{elsarticle-num-names} 
\bibliography{references}





\end{document}